\documentclass[dvips,???]{imsart}
\usepackage{graphicx}
\usepackage[numbers]{natbib}
\usepackage{amsmath,amsfonts,mathrsfs,bm,graphicx,subfigure,alg}
\usepackage{fullpage}
\usepackage[small,bf]{caption}

\newcommand{\eop}{{\hfill\vbox{\hrule height .2pt
      \hbox{\vrule width.2pt height 6pt
      \kern 4pt
      \vrule width .2pt}
      \hrule height .2pt}} \par\bigskip}
\RequirePackage[OT1]{fontenc}
\RequirePackage{amsthm,amsmath,natbib}
\RequirePackage[colorlinks,citecolor=blue,urlcolor=blue]{hyperref}
\RequirePackage{hypernat}

\startlocaldefs
\newtheorem{remark}{Remark}
\newtheorem{theorem}{Theorem}
\newtheorem{definition}{Definition}
\newtheorem{lemma}{Lemma}
\newtheorem{proposition}{Proposition}

\endlocaldefs

\begin{document}
\begin{frontmatter}

\title{Learning sparse gradients for variable selection and dimension reduction}
\runtitle{Learning sparse gradients}
\begin{aug}
\author{\fnms{Gui-Bo} \snm{Ye}
\ead[label=e1]{yeg@uci.edu}}
\and
\author{\fnms{Xiaohui} \snm{Xie}
\ead[label=e2]{xhx@ics.uci.edu}}
\address{Department of Computer Science, University of California, Irvine\\
Institute for Genomics and Bioinformatics, University of California, Irvine\\
\printead{e1,e2}}

\affiliation{University of California Irvine}
\end{aug}
\begin{abstract}
Variable selection and dimension reduction are two commonly adopted approaches for high-dimensional data analysis, but have traditionally been treated separately. Here we propose an integrated approach, called sparse gradient learning (SGL),  for  variable selection and dimension reduction via learning the gradients of the prediction function directly from samples. By imposing a sparsity constraint on the gradients, variable selection is achieved by selecting variables corresponding to non-zero partial derivatives, and effective dimensions are extracted based on the eigenvectors of the derived sparse empirical gradient covariance matrix.
An error analysis is given for the convergence of the estimated gradients to the true ones in both the Euclidean and the manifold setting. We also develop an efficient forward-backward splitting algorithm to solve the SGL problem, making the framework practically scalable for medium or large datasets. The utility of SGL  for variable selection and feature extraction is explicitly given and illustrated on artificial data as well as real-world examples. The main advantages of our method include variable selection for both linear and nonlinear predictions, effective dimension reduction with sparse loadings, and an efficient algorithm for large $p$, small $n$ problems.
\end{abstract}
\begin{keyword}[class=AMS]
\kwd[Applied statistics ]{97K80}
\kwd[; general nonlinear regression ]{62J02}
\kwd[; computational learning theory ]{68Q32}
\end{keyword}

\begin{keyword}
\kwd{Gradient learning}
\kwd{variable selection }
\kwd{effective dimension reduction}
\kwd{forward-backward splitting}
\end{keyword}
\end{frontmatter}

\section{Introduction}\label{Sec Introduction}
Datasets with many variables have become increasingly common in biological and physical sciences. In biology, it is nowadays a common practice to measure the expression values of tens of thousands of genes, genotypes of millions of SNPs, or epigenetic modifications at tens of millions of DNA sites in one single experiment. Variable selection and dimension reduction are increasingly viewed as a necessary step in dealing with these high-dimensional data.

Variable selection aims at selecting a subset of variables most relevant for predicting responses. Many algorithms have been proposed for variable selection \citep{GE:JMLR:2003}. They typically fall into two categories: Feature Ranking and Subset Selection. Feature Ranking scores each variable according to a metric, derived from various correlation or information theoretic criteria \citep{GE:JMLR:2003,Weston:JMLR:2003,Dhillon:JMLR:2003}, and eliminates variables below a threshold score. Because Feature Ranking methods select
variables based on individual prediction power, they are ineffective in selecting a subset of variables that are marginally weak but in combination strong in prediction. Subset Selection aims to overcome this drawback by considering and evaluating the prediction power of a subset of variables as a group. One popular approach to subset selection is based on direct object optimization, which formalizes an objective function of variable selection and selects variables by solving an optimization problem.
The objective function often consists
of two terms: a {\it data
fitting term} accounting for prediction accuracy, and a {\it regularization term} controlling the number of selected variables.
LASSO proposed by \citep{Tib:JRSS:1996} and elastic net  by \citep{ZH:JRSSB:2005} are
two examples of this type of approach. The two methods are widely used because of their implementation efficiency \citep{EHJT:AS:2004,ZH:JRSSB:2005} and the ability of performing simultaneous variable selection and prediction, however, a linear prediction model is assumed by both methods. The component smoothing and selection operator (COSSO) proposed in \citep{LZ:AS:2006} try to overcome this shortcoming by using a functional LASSO penalty. However, COSSO is based on the framework of smoothing spline ANOVA which makes it impossible to
deal with high dimensional data.

Dimension reduction is another commonly adopted approach in dealing with high-dimensional data. Rooting in dimension reduction is the common belief that many real-world high-dimensional data are concentrated on a low-dimensional manifold embedded in the underlying Euclidean space. Therefore mapping the high-dimensional data into the low-dimensional manifold should be able to improve prediction accuracy, to help visualize the data, and to construct better statistical models. A number of dimension reduction methods have been proposed, ranging from principle component analysis to manifold learning for non-linear settings \citep{BN:NC:2003,ZHT:JCGS:2006,Machkey:NIPS:2008,RS:Science:2000,TSL:Science:2000,DG:PNAS:2003}. However, most of these dimension reduction methods are unsupervised, and therefore are likely suboptimal with respect to predicting responses.
In supervised settings, most recent work focuses on finding a
subspace $\mathcal{S}$ such that the projection of the high dimensional data $\mathbf{x}$ onto $\mathcal{S}$
captures the statistical dependency of the response $y$ on $\mathbf{x}$. The space
$\mathcal{S}$ is called {\em effective dimension reduction} (EDR) space \citep{XTLZ:JRSS:2002}.

Several methods have been proposed to identify EDR space. The research goes back to sliced inverse regression (SIR) proposed by Li  \citep{Li:JASA:1991}, where the covariance matrix of the inverse regression is explored for dimension reduction. The main idea is that if
the conditional distribution $\rho(y|\mathbf{x})$ concentrates on a subspace $\mathcal{S}$, then the
inverse regression $E(\mathbf{x}|y)$ should lie in that same subspace. However, SIR imposes specific modeling
assumptions on the conditional distribution $\rho(y|\mathbf{x})$ or the regression $E(y|\mathbf{x})$.
These assumptions hold in particular if the distribution of $\mathbf{x}$ is elliptic.
In practice, however, we do not necessarily expect that  $\mathbf{x}$ will follow an elliptic distribution,
nor is it easy to assess departures from ellipticity in a high-dimensional setting. A further limitation of SIR is that it yields only a one-dimensional subspace for binary classifications. Other reverse regression based methods, including principal
Hessian directions (pHd \citep{Li:JASA:1992}), sliced average variance estimation (SAVE \citep{CY:ANZJS:2001})
and contour regression \citep{LZC:AS:2005}, have been proposed, but they have similar limitations. To address these limitations,
 Xia et al. \citep{XTLZ:JRSS:2002} proposed a method called the (conditional) minimum average variance estimation (MAVE) to estimate the EDR directions. The assumption underlying MAVE is quite weak and only a semiparametric model is used.
Under the semiparametric model, conditional covariance is estimated by linear smoothing and EDR directions
are then estimated by minimizing the derived conditional covariance estimation. In addition, a simple outer
product gradient (OPG) estimator is proposed as an initial estimator. Other related approaches include
methods that estimate the derivative of the regression function \citep{HJS:AS:2001,Samarov:JASA:1993}.
Recently, Fukumizu et al. \citep{FBJ:AS:2009}
proposed a new methodology which derives EDR directly from a formulation of EDR in terms of the conditional
independence of  $\mathbf{x}$ from the response $y$, given the projection of $\mathbf{x}$ on the
EDR space. The resulting estimator is shown to be consistent under weak conditions. However,
all these EDR methods can not be directly applied to the large $p$, small $n$ case,
where $p$ is the dimension of the underlying Euclidean space in which the data lies,
and $n$ is the number of samples. To deal with the large $p$, small $n$ case, Mukherjee and co-workers  \citep{MZH:JMLR:2006,MW:JMLR:2006} introduced a gradient learning method (which will be referred to as GL)  for estimating EDR by introducing a Tikhonov regularization term on the gradient functions. The EDR directions were estimated using the eigenvectors of the empirical gradient covariance matrix.

Although both variable selection and dimension reduction offer valuable tools for statistical inference in high-dimensional space and have been prominently researched, few methods are available for combining them into a single framework where variable selection and dimensional reduction can be done. One notable exception is the sparse principle component analysis (SPCA), which produces modified principle components with sparse loadings \citep{ZHT:JCGS:2006}. However, SPCA is mainly used for unsupervised linear dimension reduction, our focus here is the variable selection and dimension reduction in supervised and potentially nonlinear settings.  To motivate the reason why a combined approach might be interesting in a supervised setting, consider a microarray gene expression data measured in both normal and tumor samples. Out of $20,000$ genes measured in microarray, only a small number of genes (e.g. oncogenes) are likely responsible for gene expression changes in tumor cells. Variable selection chooses more relevant genes and  dimension reduction further extracts features based on the subset of selected genes. Taking a combined approach could potentially improve prediction accuracy by removing irrelevant noisy variables. Additionally, by focusing on a small number of most relevant genes and extracting features among them, it could also provide a more interpretable and manageable model regarding genes and biological pathways involved in the carcinogenesis.

In this article, we extend the gradient learning framework introduced by Mukherjee and co-workers  \citep{MZH:JMLR:2006,MW:JMLR:2006}, and propose a sparse gradient learning approach (SGL) for integrated variable selection and dimension reduction in a supervised setting.  The method adopts a direct object optimization approach to learn the gradient of the underlying prediction function with respect to variables, and imposes a regularization term to control the sparsity of the gradient.  The gradient of the prediction function provides a natural interpretation of the geometric structure of the data \citep{GWBV:ML:2002,MZH:JMLR:2006,MW:JMLR:2006,MWZ:Bernoulli:2009}. If a variable is irrelevant to the prediction function, the partial derivative with
respect to that variable is zero. Moreover, for non-zeros partial derivatives, the larger the norm of the partial derivative with respect to a variable is, the more important the corresponding variable is likely to be for prediction. Thus the norms of partial derivatives give us a criterion for the importance of each variable and can be used for variable selection. Motivated by LASSO, we encourage the sparsity of the gradient by adding a $\ell^1$ norm based regularization term to the objective vector function.
Variable selection is automatically achieved by selecting variables with non-zero partial derivatives.
The sparse empirical gradient covariance matrix (S-EGCM) constructed based on the learned sparse gradient reflects the variance of the data conditioned on the response variable. The eigenvectors of S-EGCM are then used to construct the EDR directions. A major innovation of our approach is that the variable selection and dimension reduction are achieved within a single framework. The features constructed by the eigenvectors of S-EGCM are sparse with non-zero entries corresponding only to selected variables.

The rest of this paper is organized as follows. In section \ref{sec
sparse gradient}, we describe the sparse gradient learning algorithm for regression,
where an automatic variable selection scheme is integrated. The
derived sparse gradient is an approximation of the true gradient of
regression function under certain conditions, which we give in subsection
\ref{sec error analysis} and their proofs are delayed in Section \ref{sec convergence analysis}.
We describe variable selection and feature construction using the learned sparse gradients in
subsection \ref{sec feature}. As our proposed algorithm is an
infinite dimensional minimization problem, it can not be solved
directly. We provide an efficient implementation for solving it in
section \ref{sec implementation}. In subsection \ref{sec
discretization}, we give a representer theorem, which transfer the
infinite dimensional sparse gradient learning problem to a finite
dimensional one. In subsection \ref{sec algorithms}, we solve
the transferred finite dimensional minimization problem by a forward-backward splitting algorithm. In section \ref{sec sparse gradient for classification},
we generalize the sparse gradient learning algorithm to a classification setting.
We illustrate the effectiveness of our gradient-based variable selection and feature extraction
approach in section \ref{sec experiment} using both simulated and real-world examples.

\section{Sparse gradient learning for regression}\label{sec sparse gradient}

\subsection{Basic definitions}\label{sec:definition}
Let $y$ and $\mathbf{x}$ be respectively $\mathbb{R}$-valued and $\mathbb{R}^p$-valued random variables. The problem of regression is to estimate the regression function $f_\rho(\mathbf{x})=\mathbb{E}(y|\mathbf{x})$ from a set of observations $\mathcal{Z} :=\{(\mathbf{x}_i,y_i)\}_{i=1}^n$,
where $\mathbf{x}_i:=(x_i^1,\ldots,x_i^p)^T\in \mathbb{R}^p$ is an
input, and $y_i\in\mathbb{R}$ is the corresponding output.

We assume the data are drawn i.i.d. from a joint distribution $\rho(\mathbf{x},y)$, and the response variable $y$ depends only on a few directions in $\mathbb{R}^p$ as follows
\begin{equation}\label{edr_assumption}
y=f_\rho(\mathbf{x})+\epsilon=g(b_1^T\mathbf{x},\ldots,b_r^T\mathbf{x})+\epsilon,
\end{equation}
where $\epsilon$ is the noise, $B=(b_1,\ldots,b_r)$ is a $p\times r$ orthogonal matrix with $r<p$, and $E(\epsilon|\mathbf{x})=0$ almost surely. We call the $r$ dimensional subspace spanned by $\{b_i\}_{i=1}^r$
the effective dimension reduction (EDR) space \citep{XTLZ:JRSS:2002}. For high-dimensional data, we further
assume that $B$ is a sparse matrix with many rows being zero vectors, i.e. the regression function depends only on a subset of variables in $\mathbf{x}$.

Suppose the regression function $f_\rho(\mathbf{x})$ is smooth.  The gradient of $f_\rho$ with respect to variables is
\begin{equation}
\nabla
f_\rho:=\left(\frac{\partial f_\rho}{\partial x^1},\ldots, \frac{\partial
f_\rho}{\partial x^p}\right)^T.
\end{equation}
A quantity of particular interest is the gradient outer product matrix $G=(G_{ij})$, a $p \times p$ matrix with elements
\begin{equation}\label{eq:gradient-outer-product}
G_{ij} := \left< \frac{\partial f_\rho}{\partial x^i},\ \frac{\partial f_\rho}{\partial x^j}\right>_{L^2_{\rho_X}},
\end{equation}
where $\rho_X$ is the marginal distribution of $\mathbf{x}$. As pointed out by Li \citep{Li:JASA:1991} and Xia et al. \citep{XTLZ:JRSS:2002}, under the assumption of the model in Eq.\ (\ref{edr_assumption}), the gradient outer product matrix $G$ is at most of rank $r$, and the EDR spaces are spanned by the eigenvectors corresponding to non-zero eigenvalues of $G$.
This observation has motivated the development of gradient-based methods for inferring the EDR directions \citep{XTLZ:JRSS:2002,MZH:JMLR:2006,MW:JMLR:2006}, and also forms the basis of our approach.

\subsection{Regularization framework for sparse gradient learning}
The optimization framework for sparse gradient learning includes a data fitting term and a regularization term.
We first describe the data fitting term. Given a set of observations $\mathcal{Z}$, a commonly used data fitting term for regression
is the mean square error
$\frac{1}{n}\sum_{i=1}^n (y_i - f_\rho({\bf x}_i))^2$.
However, because our primary goal is to estimate the gradient of $f_\rho$, we use the first order Taylor expansion to approximate $f_\rho$ by
$f_\rho({\bf x})\approx f_\rho({\bf x}_0) + \nabla f_\rho({\bf x}_0) \cdot ({\bf x}-{\bf x}_0)$. When ${\bf x}_j$ is close to ${\bf x}_i$, $f_\rho({\bf x}_j)\approx y_i + \nabla f_\rho({\bf x}_i) \cdot ({\bf x}_j-{\bf x}_i)$. Define $\vec{f}:=(f^1,\ldots,f^p)$,
where $f^j=\partial f_\rho/\partial x^j$
for $j=1,\ldots,p$. The mean square error used in our algorithm is
\begin{equation}\label{data fidelity term}
\mathcal{E}_\mathcal{Z}(\vec{f})=\frac{1}{n^2}\sum_{i,j=1}^n
\omega_{i,j}^s\big(y_i-y_j+\vec{f}(\mathbf{x}_i)\cdot(\mathbf{x}_j-\mathbf{x}_i)\big)^2
\end{equation}
considering Taylor expansion between all pairs of observations. Here $\omega_{i,j}^s$ is a weight function that ensures the locality of the approximation, i.e. $\omega_{i,j}^s\to 0$ when $\|{\bf x}_i -{\bf x}_j\|$ is large. We can use, for example, the Gaussian with standard deviation $s$ as a weight function. Let
$\omega^s(\mathbf{x})=\exp\{-\frac{\|\mathbf{x}\|^2}{2s^2}\}$.
Then the weights are given by
\begin{equation}\label{eq weight}
\omega_{i,j}^s=\omega^s(\mathbf{x}_j-\mathbf{x}_i)=\exp\left\{-\frac{\|\mathbf{x}_j-\mathbf{x}_i\|^2}{2s^2}\right\},
\end{equation}
for all $i,j=1,\cdots,n$, with parameter $s$ controlling the bandwidth of the weight function. In this paper, we view $s$ as a parameter and is fixed in implementing our algorithm, although it is possible to tune $s$ using a greedy algorithm as RODEO in \citep{LL:AS:2008}.

At first glance, this data fitting term might not appear very meaningful for high-dimensional
data as samples are typically distributed sparsely on a high dimensional space.
However, the term can also be explained in the manifold setting \citep{MWZ:Bernoulli:2009}, in which case the approximation is well defined as long as the data lying in the low dimensional manifold are relative dense. More specifically, assume $X$ is  a $d$-dimensional connected compact $C^\infty$ submanifold of $\mathbb{R}^p$ which is isometrically embedded. In particular, we know that $X$ is a metric space with the metric $d_X$ and the inclusion map $\Phi:(X,d_X)\mapsto (\mathbb{R}^p,\|\cdot\|_2)$ is well defined and continuous (actually it is $C^\infty$).
Note that the empirical data $\{\mathbf{x}_i\}_{i=1}^n$ are given in the Euclidean space $\mathbb{R}^p$ which are images of the points $\{\mathbf{q}_i\}_{i=1}^n\subset X$ under $\Phi:\mathbf{x}_i=\Phi(\mathbf{q}_i).$ Then this data fitting term \eqref{data fidelity term} can be explained in the manifold setting. From the first  order Taylor expansion,  when $\mathbf{q}_i$ and $\mathbf{q}_j$ are close enough, we can expect that
$y_j\approx y_i+\langle \nabla_Xf_\rho(\mathbf{q}_i),v_{ij}\rangle_{\mathbf{q}_i}$, where $v_{ij}\in T_{\mathbf{q}_i}X$ is the tangent vector such that $\mathbf{q}_j=\exp_{\mathbf{q}_i}(v_{ij})$. However, $v_{ij}$ is not easy to compute, we would like to represent the term $\langle\nabla_Xf_\rho(\mathbf{q}_i),v_{ij}\rangle_{\mathbf{q}_i}$ in the Euclidean space $\mathbb{R}^p$. Suppose $\mathbf{x}=\Phi(\mathbf{q})$ and $\xi=\Phi(\exp_{\mathbf{q}}(v))$ for $\mathbf{q}\in X$ and $v\in T_\mathbf{q}X$. Since $\Phi$ is  an isometric embedding, i.e. $d\Phi_\mathbf{q}:T_\mathbf{q}X\mapsto T_\mathbf{x}\mathbb{R}^p\cong\mathbb{R}^p$ is an isometry for every $\mathbf{q}\in X,$ the following holds
$$
\langle \nabla_Xf(\mathbf{q}),v\rangle_\mathbf{q}=\langle d\Phi_\mathbf{q}(\nabla_Xf(\mathbf{q})),d\Phi_\mathbf{q}(v)\rangle_{\mathbb{R}^p},
$$
where $d\Phi_\mathbf{q}(v)\approx \phi(\exp_\mathbf{q}(v))-\phi(\mathbf{q})=\xi-\mathbf{x}$ for $v\approx 0.$
Applying these relations to the observations $\mathcal{Z}=\{(\mathbf{x}_i,y_i)\}_{i=1}^n$ and denote $\vec{f}=d\Phi(\nabla_Xf)$
yields
\begin{equation}\label{data fidelity term manifold}
\mathcal{E}_\mathcal{Z}(\vec{f})=\frac{1}{n^2}\sum_{i,j=1}^n
\omega_{i,j}^s\big(y_i-y_j+\vec{f}(\mathbf{x}_i)\cdot(\mathbf{x}_j-\mathbf{x}_i)\big)^2.
\end{equation}
This is exactly the same as the one in the Euclidean setting.

Now we turn to  the regularization term on $\nabla{f}_\rho$. As discussed above, we impose a sparsity constraint on the gradient vector $\vec{f}$.
The motivation for the sparse constraint is based on the following two considerations: 1) Since most variables are assumed to be irrelevant for prediction, we expect the partial derivatives of $f_\rho$ with respect to these
variables should be zero; and 2) If  variable $x^j$ is important for prediction, we expect  the function $f_\rho$ should show significant variation
along $x^j$, and as such the norm of $\frac{\partial f_\rho}{\partial x^j}$
should be large.  Thus we will impose the sparsity constraint on the vector  $(\|\frac{\partial f_\rho}{\partial
x^1}\|,\ldots,\|\frac{\partial f_\rho}{\partial x^p}\|)^T\in
\mathbb{R}^p$, where $\|\cdot\|$ is a function norm, to regularize the number of non-zeros entries in the vector.

In this work, we specify  the function norm $\|\cdot\|$ to be
$\|\cdot\|_{\mathcal{K}}$, the norm in reproducing kernel Hilbert
space (RKHS) $\mathbb{H}_{\mathcal{K}}$ associated with a Mercer
kernel $\mathcal{K}(\cdot,\cdot)$ (see \citep{Aron} and Section
\ref{sec discretization}).
The sparsity constraint on the gradient norm vector implies
that the $\ell_0$ norm of the
vector $(\|f^1\|_{\mathcal{K}},\ldots, \|f^p\|_{\mathcal{K}})^T$
should be small. However, because the $\ell_0$ norm is difficult to work with during optimization, we instead use the
$\ell_1$ norm of the vector \citep{Don:IEEE:95,DDD:CPAM:04,Don:IEEE:2006} as our regularization term
\begin{equation}\label{regularization term}
\Omega(\vec{f}):=\lambda\sum_{j=1}^p\|f^j\|_{\mathcal{K}},
\end{equation}
where $\lambda$ is a sparsity regularization parameter. This functional LASSO penalty has been used in \cite{LZ:AS:2006} as COSSO penalty.
However, our component here is quite different from theirs, which makes our algorithm  useful for high dimensional problems.

The norm $\|\cdot\|_{\mathcal{K}}$ is widely used in statistical inference and
machine learning (see \citep{Vapnik:book:1998}). It can
ensure each approximated partial derivative
$f^j\in\mathbb{H}_{\mathcal{K}}$, which in turn imposes some
regularity on each partial derivative. It is possible to replace
the hypothesis space $\mathbb{H}_\mathcal{K}^p$ for the vector
$\vec{f}$ in \eqref{regularization term} by some other space of
vector-valued functions \citep{MP:NC:2005} in order to learn the
gradients.

Combining the data fidelity term \eqref{data fidelity term} and the
regularization term \eqref{regularization term}, we propose the
following optimization framework, which will be referred as \emph{sparse gradient learning},  to learn $\nabla{f}_\rho$
\begin{equation}\label{gradient learning algorithm}
\vec{f}_{\mathcal{Z}}:=\arg\min_{\vec{f}\in {\mathbb
H}_{\mathcal{K}}^p}\
\frac{1}{n^2}\sum_{i,j=1}^n
\omega_{i,j}^s\big(y_i-y_j+\vec{f}(\mathbf{x}_i)\cdot(\mathbf{x}_j-\mathbf{x}_i)\big)^2 +
\lambda\sum_{j=1}^p\|f^j\|_{\mathcal{K}}
.
\end{equation}

A key difference between our framework and the one in \citep{MZH:JMLR:2006} is that our regularization is based on $\ell_1$ norm, while the one in \citep{MZH:JMLR:2006} is based on ridge regularization. The difference may appear minor, but makes a significant impact on the estimated $\nabla f_\rho$. In particular, $\nabla f_\rho$ derived from Eq.\ (\ref{gradient learning algorithm}) is sparse with many components potentially being zero functions, in contrast to the one derived from \citep{MZH:JMLR:2006}, which is comprised of all non-zero functions.
The sparsity property is desirable for two primary reasons: 1) In most high-dimensional real-world data, the response variable is known to depend only on  a subset of the variables. Imposing sparsity constraints can help eliminate noisy variables and thus improve the accuracy for inferring the EDR directions; 2) The resulting gradient vector provides a way to automatically select and rank relevant variables.

\begin{remark}
The OPG method introduced by Xia et al. \citep{XTLZ:JRSS:2002} to learn EDR directions can be viewed as a special case of the sparse gradient learning, corresponding to the case of setting $K(x,y)=\delta_{x,y}$ and $\lambda=0$ in Eq.\ (\ref{gradient learning algorithm}). Thus the sparse gradient learning can be viewed as an extension of learning gradient vectors only at observed points by OPG to a vector function of gradient over the entire space. Note that OPG cannot be directly applied to the data with $p> n$ since the problem is then underdetermined. Imposing a regularization term as in  Eq.\ (\ref{gradient learning algorithm}) removes such a limitation.
\end{remark}

\begin{remark}
The sparse gradient learning reduces to a special case that is approximately LASSO \citep{Tib:JRSS:1996} if we choose $K(x,y)=\delta_{x,y}$ and additionally require $\vec{f}(\mathbf{x}_i)$ to be invariant for different $i$ (i.e. linearity assumption). Note that LASSO assumes the regression function is linear, which can be problematic for variable selection when the prediction function is nonlinear \citep{EHJT:AS:2004}.  The sparse gradient learning makes no linearity assumption, and can thus be viewed as an extension of LASSO for variable selection with nonlinear prediction functions.
\end{remark}

\begin{remark}
A related framework is to learn the regression function directly, but impose constraints on the sparsity of the
 gradient as follows
 \begin{equation}
\min_{f\in
\mathbb{H}_\mathcal{K}}\frac{1}{n}\sum_{i=1}^n(f(\mathbf{x}_i)-y_i)^2+\lambda\sum_{i=1}^p
\|\frac{\partial f}{\partial x^i}\|_\mathcal{K}.
\end{equation}
This framework is however difficult to solve because the regularization term $\sum_{i=1}^p \|\frac{\partial
f}{\partial x^i}\|_\mathcal{K}$ is both nonsmooth and inseparable, and the representer theorem introduced later to solve Eq.\ (\ref{gradient learning algorithm}) cannot be applied here. Note that our primary goal is to select variables and identify
the EDR directions. Thus we focus on learning gradient functions rather than the regression function itself.
\end{remark}

\subsection{Error analysis}\label{sec error analysis}

Next we investigate the statistical performance of the sparse gradient learning with a Gaussian weight in Eq. \eqref{eq weight}. Assume that the data
$\mathcal{Z}=\{(\mathbf{x}_i,y_i)\}_{i=1}^n$ are i.i.d drawn from a
joint distribution $\rho$, which can be divided into a marginal
distribution $\rho_X$ and a conditional distribution $\rho(y|\mathbf{x})$.
Denote $f_\rho$ to be the regression function given by
$$f_\rho(x)=\int_Yyd\rho(y|\mathbf{x}).$$

We show that under certain conditions,
$\vec{f}_\mathcal{Z}\rightarrow \nabla f_\rho$ as
$n\rightarrow \infty$ for suitable choices of the parameters
$\lambda$ and $s$ that go to zero as $n \to \infty$.
In order to derive the learning
rate for the algorithm, some regularity conditions on both the
marginal distribution and $\nabla f_\rho$ are required.

Denote $\partial X$ be the boundary of $X$ and  $d(\mathbf{x},\partial X) (\mathbf{x}\in X)$ be the shortest Euclidean distance from
$\mathbf{x}$ to $\partial X$, i.e, $d(\mathbf{x},\partial X)=\inf_{\mathbf{y}\in \partial X}d(\mathbf{x},\mathbf{y})$.

\begin{theorem}\label{thm error rates}
Suppose  the data $\mathcal{Z}=\{(\mathbf{x}_i,y_i)\}_{i=1}^n$ are   i.i.d drawn
from a joint distribution $\rho$ and  $y_i\leq M$ for all $i$ for a
positive constant $M$. Assume that for some constants $c_\rho>0$ and
$0<\theta\leq 1$, the marginal distribution $\rho_X$ satisfies
\begin{equation}\label{inq marginal distribution one}
\rho_X(\{\mathbf{x}\in X:d(\mathbf{x},\partial X)\}<t)\leq c_\rho t
\end{equation}
and the density $p(\mathbf{x})$ of $\rho_X$ satisfies
\begin{equation}\label{inq density one}
\sup_{\mathbf{x}\in X}p(\mathbf{x})\leq c_\rho \ \hbox{and} \ |p(\mathbf{x})-p(\mathbf{u})|\leq c_\rho
|\mathbf{x}-\mathbf{u}|^\theta,\ \forall \mathbf{u},\mathbf{x}\in X.
\end{equation}
Let $\vec{f}_\mathcal{Z}$ be the estimated gradient function given by Eq.\
\eqref{gradient learning algorithm} and $\nabla f_\rho$ be the true
gradient of the regression function $f_\rho$. Suppose that $\mathcal{K}\in
C^2$ and $\nabla f_\rho\in \mathbb{H}_\mathcal{K}^p$. Choose
$\lambda=\lambda(n)=n^{-\frac{\theta}{p+2+2\theta}}$ and
$s=s(n)=n^{-\frac{1}{2(p+2+2\theta)}}$. Then there exists a constant
$C>0$ such that for any $0< \eta\leq1$ with confidence $1-\eta$
\begin{equation}\label{eq learning rate}
\|\vec{f}_\mathcal{Z}-\nabla f_\rho\|_{L^2_{\rho_X}}\leq
C
\log\frac{4}{\eta}\left(\frac{1}{n}\right)^{\frac{\theta}{4(p+2+2\theta)}}.
\end{equation}
\end{theorem}
Condition \eqref{inq density one} means the density of the marginal
distribution is H\"{o}lder continuous with exponent $\theta$. Condition \eqref{inq
marginal distribution} specifies behavior of $\rho_X$ near the
boundary $\partial X$ of $X$. Both are common assumptions for error analysis. When  the boundary $\partial X$ is  piecewise
smooth, Eq.\ \eqref{inq density one} implies Eq.\ \eqref{inq marginal
distribution}. Here we want to emphasize that our terminology sparse gradient for the derived $\vec{f}_\mathcal{Z}$ comes from this approximation property. Since we treat each component of the gradient separately in our estimation algorithm, $\vec{f}_\mathcal{Z}$ does not necessarily satisfy the gradient constraint $\frac{\partial ^2 f}{\partial x^i\partial x^j}=\frac{\partial ^2 f}{\partial x^j\partial x^i}$ for all $i$ and $j$. However, we note that it is possible to add these constraints  explicitly into the convex optimization framework that we will describe later.

The convergence rate in Eq.\ \eqref{eq learning rate} can be greatly improved if we assume that the data are lying in or near a low dimensional manifold \citep{YeZh:ACM:2008,MWZ:Bernoulli:2009}. In this case, the learning rate in the exponent of $1/n$ depends only on the dimension of the manifold, not the actual dimension of the Euclidean space.

Denote $d_X$ be the metric on $X$ and $dV$ be the Riemannian volume  measure of $M$. Let $\partial X$ be the boundary of $X$ and $d_X(\mathbf{x},\partial X) (\mathbf{x}\in X)$ be the shortest  distance from
$\mathbf{x}$ to $\partial X$ on the manifold $X$. Denote $(d\Phi)^*_\mathbf{q}$ is the dual of $d\Phi_\mathbf{q}$ and $(d\Phi)^*$
maps a $p$-dimensional vector valued function $\vec{f}$ to a vector field with $(d\Phi)^*\vec{f}(\mathbf{q})=(d\Phi)_\mathbf{q}^*(\vec{f}(\bf{q}))$ \cite{Carmo:book:1992}.
\begin{theorem}\label{Theorem convergence rate manifold}
Let $X$ be a connected compact $C^\infty$ submanifold of $\mathbb{R}^p$ which is isometrically embedded and of dimension $d$.
Suppose  the data $\mathcal{Z}=\{(\mathbf{x}_i,y_i)\}_{i=1}^n$ are   i.i.d drawn
from a joint distribution $\rho$ defined on $X\times Y$ and there exists a
positive constant $M$ such that $y_i\leq M$ for all $i$. Assume that for some constants $c_\rho>0$ and
$0<\theta\leq 1$, the marginal distribution $\rho_X$ satisfies
\begin{equation}\label{inq marginal distribution}
\rho_X(\{\mathbf{x}\in X:d_X(\mathbf{x},\partial X)\}<t)\leq c_\rho t
\end{equation}
and the density $p(\mathbf{x})=\frac{d\rho_X(\mathbf{x})}{dV}$ exists and  satisfies
\begin{equation}\label{inq density}
\sup_{\mathbf{x}\in X}p(\mathbf{x})\leq c_\rho \ \hbox{and} \ |p(\mathbf{x})-p(\mathbf{u})|\leq c_\rho
d_X(\mathbf{x},\mathbf{u})^\theta,\ \forall \mathbf{u},\mathbf{x}\in X.
\end{equation}
Let $\vec{f}_\mathcal{Z}$ be the estimated gradient function given by Eq.\
\eqref{gradient learning algorithm} and $\nabla_X f_\rho$ be the true
gradient of the regression function $f_\rho$. Suppose that $\mathcal{K}\in
C^2(X\times X)$, $f_\rho\in C^2(X)$ and $d\Phi(\nabla_X f_\rho)\in \mathbb{H}_\mathcal{K}^p$. Choose
$\lambda=\lambda(n)=n^{-\frac{\theta}{d+2+2\theta}}$ and
$s=s(n)=n^{-\frac{1}{2(d+2+2\theta)}}$. Then there exists a constant
$C>0$ such that for any $0< \eta\leq1$ with confidence $1-\eta$
\begin{equation}\label{eq learning rate}
\|(d\Phi)^*\vec{f}_\mathcal{Z}-\nabla_X f_\rho\|_{L^2_{\rho_X}}\leq
C
\log\frac{4}{\eta}\left(\frac{1}{n}\right)^{\frac{\theta}{4(d+2+2\theta)}}.
\end{equation}
\end{theorem}
Note that the convergence rate in Theorem \ref{Theorem convergence rate manifold} is exactly the same as the one in Theorem \ref{thm error rates} except that we replaced the Euclidean dimension $p$ by the intrinsic dimension $d$.

The constraints $\nabla_X f_\rho\in \mathcal{H}^p_\mathcal{K}$ in Theorem \ref{thm error rates} and $d\Phi(\nabla_Xf_\rho)\in \mathcal{H}^p_\mathcal{K}$ are somewhat restrictive, and extension to mild conditions is possible \cite{MWZ:Bernoulli:2009}. Here we confine ourself to these conditions in order to avoid introducing more notations and conceptions.

The proof of Theorem \ref{thm error rates} and Theorem \ref{Theorem convergence rate manifold} are somewhat complicated and will be given in the Section \ref{sec convergence analysis}. The main idea behind the proof is to  simultaneously control the sample error and the approximation error; see section \ref{sec convergence analysis} for details.

\subsection{Variable selection and effective dimension reduction}\label{sec feature}

Next we describe how to do variable selection and extract EDR directions based on the learned gradient
$\vec{f}_{\mathcal{Z}}=(f_{\mathcal{Z}}^1,\ldots,f_{\mathcal{Z}}^p)^T$.

As discussed above, because of the $l_1$ norm used in the regularization term, we expect many of the entries in the gradient vector
$\vec{f}_{\mathcal{Z}}$ be zero functions.  Thus, a natural way to select variables is to identify those entries with non-zeros functions. More specifically, we select variables based on the following criterion.
\begin{definition}
Variable selection via sparse gradient learning is to select variables in the set
\begin{equation}\label{index set}
\mathcal{S}:=\{j: \|f_{\mathcal{Z}}^j \|_{\mathcal K} \neq 0,~~j=1,\ldots,p\}
\end{equation}
where $\vec{f}_{\mathcal{Z}}=(f_{\mathcal{Z}}^1,\ldots,f_{\mathcal{Z}}^p)^T$ is the estimated gradient vector.
\end{definition}

To select the EDR directions, we focus on the empirical gradient covariance matrix defined below
\begin{equation}\label{EGCM}
\Xi:=\left[\langle
f^i_{\mathcal{Z}},f^j_{\mathcal{Z}}\rangle_{\mathcal{K}}\right]_{i,j=1}^p.
\end{equation}
The inner product $\langle f_{\mathcal{Z}}^i,f_{\mathcal{Z}}^j\rangle_{\mathcal{K}}$ can be interpreted as the covariance of the gradient functions between coordinate $i$ and $j$. The larger the inner product is, the
more related the variables $x^i$ and $x^j$ are.
Given a unit vector $\mathbf{u}\in \mathbb{R}^p$, the RKHS norm of
the directional derivative $\|\mathbf{u}\cdot
\vec{f}_\mathcal{Z}\|_\mathcal{K}$ can be viewed as a measure of the
variation of the data $\mathcal{Z}$ along the direction
$\mathbf{u}$. Thus the direction $\mathbf{u}_1$ representing the largest variation in the data is the vector that maximizes $\|\mathbf{u}\cdot\vec{f}_{\mathcal{Z}}\|^2_{\mathcal{K}}$. Notice
that
$$\|\mathbf{u}\cdot\vec{f}_{\mathcal{Z}}\|^2_{\mathcal{K}}=\|
\sum_iu_if_{\mathcal{Z}}^i\|_{\mathcal{K}}^2=\sum_{i,j}u_iu_j\langle
f_{\mathcal{Z}}^i,f_{\mathcal{Z}}^j\rangle_{\mathcal{K}}=\mathbf{u}^T\Xi\mathbf{u}.$$
So $\mathbf{u}_1$ is simply the eigenvector of $\Xi$ corresponding to the
largest eigenvalue. Similarly, to construct the second most
important direction $\mathbf{u}_2$, we maximize
$\|\mathbf{u}\cdot\vec{f}_{\mathcal{Z}}\|_{\mathcal{K}}$ in the
orthogonal complementary space of $\mathrm{span}\{\mathbf{u}_1\}$.
By Courant-Fischer Minimax Theorem \citep{Golub:book}, $\mathbf{u}_2$
is the eigenvector corresponding to the second largest eigenvalue of
$\Xi$. We repeat this procedure to construct other important directions. In summary, the effective dimension reduction directions are defined according to the following criterion.

\begin{definition}
The $d$ EDR directions identified by the sparse gradient learning are the eigenvectors  $\{\mathbf{u}_1,\ldots,\mathbf{u}_d\}$ of $\Xi$ corresponding to the $d$ largest eigenvalues.
\end{definition}

As we mentioned in section \ref{sec:definition}, the EDR space is spanned by the eigenvectors of the gradient outer product matrix $G$ defined in Eq.\ (\ref{eq:gradient-outer-product}).  However, because the distribution of the data is unknown, $G$ cannot be calculated explicitly. The above definition provides a way to approximate the EDR directions based on the empirical gradient covariance matrix.

Because of the sparsity of the estimated gradient functions, matrix $\Xi$ will appear to be block sparse. Consequently, the identified EDR directions will be sparse as well with non-zeros entries only at coordinates belonging to the set $S$.  To emphasize the sparse property of both $\Xi$ and the identified EDR directions, we will refer to $\Xi$ as the sparse empirical
gradient covariance matrix (S-EGCM), and the identified EDR directions as the sparse effective dimension reduction directions (S-EDRs).

\section{Convergence Analysis}\label{sec convergence analysis}
In this section, we will give the proof of Theorem  \ref{thm error
rates} and Theorem \ref{Theorem convergence rate manifold}.

\subsection {Convergence Analysis in the Euclidean Setting} \label{sec proof of Thm error
rates}
Note that our energy functional in \eqref{gradient learning algorithm} involves an nonsmooth regularization term $\sum_i\|f^i\|_\mathcal{K}$. The method for the convergence analysis used in \cite{MZH:JMLR:2006}
can no longer be applied any more since it need explicit form of the solution which is only possible for the $\ell^2$
regularization. However, we can still simultaneously control a sample or
estimation error term and a regularization or approximation error
term which is widely used in statistical learning theory \cite{Vapnik:book:1998,MW:JMLR:2006,Zhang.AS:2004}.

\subsubsection {Comparative Analysis}
Recall the empirical error for a
vector function $\vec{f}:=(f^1,\ldots,f^p)$,
$$\mathcal{E}_\mathcal{Z}(\vec{f})=\frac{1}{n^2}\sum_{i,j=1}^n
\omega_{i,j}^s\big(y_i-y_j+\vec{f}(\mathbf{x}_i)\cdot(\mathbf{x}_j-\mathbf{x}_i)\big)^2.$$
One can similarly define the expected error
$$\mathcal{E}(\vec{f})=\int_Z\int_Z\omega^s(\mathbf{x}-\mathbf{u})(y-v+\vec{f}(\mathbf{x})(\mathbf{u}-\mathbf{x}))^2d\rho(\mathbf{x},y)d\rho(\mathbf{u},v).$$

Denote
$$\sigma_s^2=\int_X\int_Z\omega^s(\mathbf{x}-\mathbf{u})(y-f_\rho(\mathbf{x}))^2d\rho(\mathbf{x},y)d\rho_X(\mathbf{u}).$$
Then
$\mathcal{E}(\vec{f})=2\sigma_s^2+\int_X\int_X\omega(\mathbf{x}-\mathbf{u})[f_\rho(\mathbf{x})-f_\rho(\mathbf{u})
+\vec{f}(\mathbf{x})(\mathbf{u}-\mathbf{x})]^2d\rho_X(\mathbf{x})d\rho_X(\mathbf{u}).$

Note that our goal is to bound the $L_{\rho_X}^2$ differences of $\vec{f}$ and
$\nabla{f_\rho}$. We have the following comparative theorem to
bound the $L_{\rho_X}^2$ differences of $\vec{f}$ and
$\nabla{f_\rho}$ in terms of  the excess error, $\mathcal{E}(\vec{f})-2\sigma_s^2$ using the following comparative theorem.

For $r>0$, denote
$$\mathcal{F}_r=\{\vec{f}\in \mathcal{H}_\mathcal{K}^p:\sum_{i=1}^p\|f^i\|_\mathcal{K}\leq r\}.$$
\begin{theorem}\label{thm appendix one}
Assume $\rho_X$ satisfies the condition \eqref{inq marginal
distribution one} and \eqref{inq density one} and $\nabla f_\rho\in
\mathcal{H}_\mathcal{K}^p.$ For $\vec{f}\in \mathcal{F}_r$ with some
$r\geq1$, there exist a constant $C_0>0$ such that
$$\|\vec{f}-\nabla f_\rho\|_{L_{\rho_X}^2}\leq C_0(r^2s^\theta+s^{2-\theta}+\frac{1}{s^{p+2+\theta}}(\mathcal{E}(\vec{f})-2\sigma_s^2)).$$
\end{theorem}

To prove Theorem \ref{thm appendix one}, we need several lemmas
which require the notations of the following quantities. Denote
$$Q(\vec{f})=\int_X\int_X\omega(\mathbf{x}-\mathbf{u})((\vec{f}(\mathbf{x})-\nabla f_\rho(\mathbf{x}))
(\mathbf{u}-\mathbf{x}))^2d\rho_X(\mathbf{x})d\rho_X(\mathbf{u}),$$
the border set $$X_s=\{\mathbf{x}\in X:d(\mathbf{x},\partial X)>s \
\hbox{and}\ p(\mathbf{x})\geq (1+c_\rho)s^\theta\}$$ and the moments for
$0\leq q<\infty$
$$M_q=\int_{\mathbb{R}^p}e^{-\frac{\|\mathbf{t}\|^2}{2}}\|\mathbf{t}\|^qd\mathbf{t},\qquad \widetilde{M}_q
=\int_{\|\mathbf{t}\|\leq
1}e^{-\frac{\|\mathbf{t}\|^2}{2}}\|\mathbf{t}\|^qd\mathbf{t}.$$ Note
that $X_s$ is nonempty when $s$ is small enough.
\begin{lemma}\label{lemma appendix one}
Under assumptions of Theorem \ref{thm appendix one},
$$\frac{\widetilde{M}_2s^{p+2+\theta}}{p}\int_{X_s}\|\vec{f}(\mathbf{x})-\nabla f_\rho(\mathbf{x})\|^2d\rho_X(\mathbf{x})\leq Q(\vec{f})$$
\end{lemma}
\begin{proof}
For $\mathbf{x}\in X_s,$ we have $d(\mathbf{x},\partial X)>s$ and $p(\mathbf{x})\geq (1+c_\rho)s^\theta$. Thus
$\{\mathbf{u}\in X:|\mathbf{u}-\mathbf{x}|\leq s\}\subset X$ and for
$\mathbf{u}\in \{\mathbf{u}\in X:|\mathbf{u}-\mathbf{x}|\leq s\}$,
$p(\mathbf{u})=p(\mathbf{x})-(p(\mathbf{x})-p(\mathbf{u}))\geq
(1+c_\rho)s^\theta-c_\rho|\mathbf{u}-\mathbf{x}|^\theta\geq
s^\theta$. Therefore,
\begin{eqnarray}
Q(\vec{f})&\geq& \int_{X_s}\int_{\|\mathbf{x}-\mathbf{u}\|\leq
s}\omega^s(\mathbf{x}-\mathbf{u})((\vec{f}(\mathbf{x})-\nabla
f_\rho(\mathbf{x}))(\mathbf{x}-\mathbf{u}))^2p(\mathbf{u})d\mathbf{u}d\rho_X(\mathbf{x})\nonumber\\
&\geq& s^\theta \int_{X_s}\int_{\|\mathbf{x}-\mathbf{u}\|\leq
s}\omega^s(\mathbf{x}-\mathbf{u})((\vec{f}(\mathbf{x})-\nabla
f_\rho(\mathbf{x}))(\mathbf{x}-\mathbf{u}))^2d\mathbf{u}d\rho_X(\mathbf{x}).\nonumber
\end{eqnarray}
Denote the $i$-th entry of a vector $\mathbf{x}$ by $x^i$. Then
$((\vec{f}(\mathbf{x})-\nabla f_\rho(\mathbf{x}))(\mathbf{x}-\mathbf{u}))^2$
equals to
$$\sum_{i=1}^p\sum_{j=1}^p(f^i(\mathbf{x})-\frac{\partial f_\rho}{\partial
x^i}(\mathbf{x})) (f^j(\mathbf{x})-\frac{\partial f_\rho}{\partial
x^j}(\mathbf{x}))(x^i-u^i)(x^j-u^j).$$ For the case $i\neq j$, we
have
$$\int_{\|\mathbf{u}-\mathbf{x}\|\leq s}\omega^s(\mathbf{x}-\mathbf{u})(x^i-u^i)(x^j-u^j)d\mathbf{u}
=s^{p+2}\int_{\|\mathbf{t}\|\leq
1}e^{-\frac{\|\mathbf{t}\|^2}{2}}t^it^jd\mathbf{t}=0.$$ Therefore,
\begin{eqnarray}
Q(\vec{f})&\geq& s^{p+2+\theta}
\sum_{i=1}^p\int_{X_s}(f^i(\mathbf{x})-\frac{\partial
f_\rho}{\partial
x^i}(\mathbf{x}))^2d\rho_X(\mathbf{x})\int_{\|\mathbf{t}\|\leq
1}e^{-\frac{\|\mathbf{t}\|^2}{2}}(t^i)^2d\mathbf{t}\nonumber\\
&=&\frac{\widetilde{M}_2s^{p+2+\theta}}{p}\int_{X_s}\|\vec{f}(\mathbf{x})-\nabla
f_\rho(\mathbf{x})\|^2d\rho_X(\mathbf{x}),\nonumber
\end{eqnarray}
which yields the desired estimate.
\end{proof}
\begin{lemma}\label{lemma appendix two}
Under the assumption of Theorem \ref{thm appendix one}, we have
$$Q(\vec{f})\leq C_1(s^{4+p}+\mathcal{E}(\vec{f})-2\sigma_s^2),$$
where $C_1$ is a constant independent of $s$ or $\vec{f}$.
\end{lemma}
\begin{proof}
Denote $a_1=(\vec{f}(\mathbf{x})-\nabla
f_\rho(\mathbf{x}))(\mathbf{u}-\mathbf{x})$  and
$a_2=f_\rho(\mathbf{x})-f_\rho(\mathbf{u})+\nabla f_\rho
(\mathbf{x})(\mathbf{u}-\mathbf{x}).$ We have
$Q(\vec{f})=\int_X\int_X\omega^s(\mathbf{x}-\mathbf{u})(a_1)^2d\rho_X(\mathbf{x})d\rho_X(\mathbf{u})$
and
$$\mathcal{E}(\vec{f})=\int_X\int_X\omega^s(\mathbf{x}-\mathbf{u})(a_1+a_2)^2d\rho_X(\mathbf{x})d\rho_X(\mathbf{u})+2\sigma_s^2.$$
Note that $(a_1+a_2)^2\geq (a_1)^2-2\|a_1\|\|a_2\|.$ Thus
$$\mathcal{E}(\vec{f})-2\sigma_s^2\geq Q(\vec{f})-2\int_X\int_X\omega^s(\mathbf{x}-\mathbf{u})
\|a_1\|\|a_2\|d\rho_X(\mathbf{x})d\rho_X(\mathbf{u}).$$ By the fact
$\nabla f_\rho\in \mathbb{H}_{\mathcal{K}}^p$ and lemma $19$ in
\citep{MW:JMLR:2006}, there exists a constant $C_\mathcal{K}>0$
depending  on $\mathcal{K}$ and $f_\rho$ such that
$$\|a_2\|\leq C_\mathcal{K} \|\mathbf{x}-\mathbf{u}\|^2.$$
Together with the assumption $p(\mathbf{x})\leq c_\rho$, we have
\begin{eqnarray}
&&\int_X\int_X\omega^s(\mathbf{x}-\mathbf{u})\|a_1\|\|a_2\|d\rho_X(\mathbf{x})d\rho_X(\mathbf{u})\nonumber\\
&\leq&\sqrt{Q(\vec{f})}(\int_X\int_X\omega^s(\mathbf{x}-\mathbf{u})\|a_2\|^2d\rho_X(\mathbf{x})d\rho_X(\mathbf{u}))^{\frac{1}{2}}\nonumber\\
&\leq&C_\mathcal{K}\sqrt{Q(\vec{f})}(c_\rho\int_X\int_{\mathbb{R}^p}\omega^s(\mathbf{x}-\mathbf{u})\|\mathbf{x}-\mathbf{u}\|^4d
\mathbf{x}d\rho_X(\mathbf{u}))^{\frac{1}{2}}\nonumber\\
&\leq& C_\mathcal{K}\sqrt{c_\rho M_4}s^{2+p/2}\sqrt{Q(\vec{f})}.
\end{eqnarray}
Combining the above arguments, we obtain
$$Q(\vec{f})-C_\mathcal{K}\sqrt{c_\rho M_4}s^{2+p/2}\sqrt{Q(\vec{f})}\leq\mathcal{E}(\vec{f})-2\sigma_s^2.$$
This implies the conclusion with $C_1=2\max\{C_\mathcal{K}^2c_\rho
M_4,1\}.$
\end{proof}
Denote $\kappa=\sup_{x\in X}\sqrt{K(x,x)},
D=\max_{\mathbf{x},\mathbf{u}\in X}\|\mathbf{x}-\mathbf{u}\|.$

{\bf Proof of Theorem \ref{thm appendix one}.} Write
\begin{equation}\label{eq decomposition L2}
\|\vec{f}-\nabla f_\rho\|_{L^2_{\rho_X}}^2=\int_{X\backslash
X_s}\|\vec{f}(\mathbf{x})-\nabla
f_\rho(\mathbf{x})\|^2d\rho_X(\mathbf{x})
+\int_{X_s}\|\vec{f}(\mathbf{x})-\nabla
f_\rho(\mathbf{x})\|^2d\rho_X(\mathbf{x}).
\end{equation}
We have
$$\rho_X(X\backslash X_s)\leq c_\rho s+(1+c_\rho)c_\rho |X|s^\theta\leq (c_\rho+(1+c_\rho)c_\rho|X|)s^\theta,$$
where $|X|$ is the Lebesgue measure of $X$. So the first term on the
right of \eqref{eq decomposition L2} is bounded by
$$\kappa^2(r+\|\nabla f_\rho\|_\mathcal{K})^2(c_\rho+(1+c_\rho)c_\rho|X|)s^\theta.$$
By lemma \ref{lemma appendix one} and lemma \ref{lemma appendix
two}, the second term on the right hand of \eqref{eq decomposition
L2} is bounded by
$$\frac{pC_1}{\widetilde{M}_2}\frac{1}{s^{p+2+\theta}}(s^{4+p}+\mathcal{E}(\vec{f})-2\sigma_s^2).$$
Combining these two estimates finishes the proof of the  claim with
$$C_0=\kappa^2(1+\|\nabla
f_\rho\|_\mathcal{K})^2(c_\rho+(1+c_\rho)c_\rho|X|)+\frac{pC_1}{\widetilde{M}_2}.$$
This is the end of the proof.

\subsubsection{Error Decomposition}
Now we turn to bound the quantity
$\mathcal{E}(\vec{f}_\mathcal{Z})-2\sigma_s^2$. Note that unlike the standard setting of regression and classification,
$\mathcal{E}_\mathcal{Z}(\vec{f})$ and $\mathcal{E}(\vec{f})$ are
not respectively the expected and empirical mean of a random
variable. This is due to the extra $d\rho(\mathbf{u},v)$ in the
expected error term. However, since
$$E_\mathcal{Z}\mathcal{E}_\mathcal{Z}(\vec{f})=\frac{n-1}{n}\mathcal{E}({\vec{f}}),$$
$\mathcal{E}_\mathcal{Z}(\vec{f})$ and $\mathcal{E}(\vec{f})$ should
be close to each other if the empirical error concentrates with $n$
increasing. Thus, we can still decompose $\mathcal{E}(\vec{f}_\mathcal{Z})-2\sigma_s^2$ into a sample error term and an approximation error term.

Note that $\Omega(\vec{f})=\lambda\sum_i\|f^i\|_\mathcal{K}$ with
$\vec{f}=(f^1,\ldots,f^p),$ so the minimizer of $\mathcal{E}(\vec{f})+\Omega(\vec{f})$ in $\mathbb{H}_\mathcal{K}^p$
depends on $\lambda$. Let
\begin{equation}\label{eq regularization function}
\vec{f}_\lambda=\arg\min_{\vec{f}\in \mathbb{H}_\mathcal{K}^p}\{\mathcal{E}(\vec{f})+
\Omega(\vec{f})\}.
\end{equation}

By a standard decomposition procedure, we have the following result.
\begin{proposition}\label{prop appendix one}Let $$\varphi(\mathcal{Z})=(\mathcal{E}(\vec{f}_{\mathcal{Z}})-\mathcal{E}_\mathcal{Z}(\vec{f}_\mathcal{Z}))
+(\mathcal{E}_\mathcal{Z}(\vec{f}_\lambda))-\mathcal{E}(\vec{f}_\lambda))$$
and $$\mathcal{A}(\lambda)=\inf_{\vec{f}\in
\mathbb{H}_\mathcal{K}^p}\{\mathcal{E}(\vec{f})-2\sigma_s^2+\Omega(\vec{f})\}.$$
Then, we have
$$\mathcal{E}(\vec{f}_\mathcal{Z})-2\sigma_s^2\leq\mathcal{E}(\vec{f}_\mathcal{Z})-2\sigma_s^2+ \Omega(\vec{f}_\mathcal{Z})\leq \varphi(\mathcal{Z})+\mathcal{A}(\lambda)$$
\end{proposition}
The quantity $\varphi(\mathcal{Z})$ is called the sample error and $\mathcal{A}(\lambda)$ is the approximation error.
\subsubsection{Sample Error Estimation}
Note that the sample error $\varphi(\mathcal{Z})$
can be bounded by controlling
$$S(\mathcal{Z},r):=\sup_{\vec{f}\in \mathcal{F}_r}|\mathcal{E}_\mathcal{Z}(\vec{f})-\mathcal{E}(\vec{f})|.$$
In fact, if both $\vec{f}_\mathcal{Z}$ and $\vec{f}_\lambda$ are in
$\mathcal{F}_r$ for some $r>0$, then
\begin{equation}\label{inq sample error bound one}
\varphi(\mathcal{Z})\leq 2S(\mathcal{Z},r).
\end{equation}
We use McDiarmid's inequality in \citep{McDiarmid:SC:1989} to bound
$S(\mathcal{Z},r)$.
\begin{lemma}\label{lemma appendix four}
For every $r>0$,
$$\hbox{Prob}\{|S(\mathcal{Z},r)-ES(\mathcal{Z},r)|\geq \epsilon\}\leq 2\exp\left(-\frac{n\epsilon^2}{32(M+\kappa Dr)^4}\right).$$
\end{lemma}
\begin{proof}
Let $(\mathbf{x}^\prime,y^\prime)$ be a sample i.i.d drawn from the
distribution $\rho(\mathbf{x},y)$. Denote by $\mathcal{Z}_i^\prime$
the sample which coincides with $\mathcal{Z}$ except that the $i$-th
entry $(\mathbf{x}_i,y_i)$ is replaced by
$(\mathbf{x}^\prime,y^\prime)$. It is easy to verify that
\begin{eqnarray}
S(\mathcal{Z},r)-S(\mathcal{Z}_i^\prime,r)&=&\sup_{\vec{f}\in
\mathcal{F}_r}|\mathcal{E}_\mathcal{Z}(\vec{f})-\mathcal{E}(\vec{f})|-\sup_{\vec{f}\in
\mathcal{F}_r}|\mathcal{E}_{\mathcal{Z}_i^\prime}(\vec{f})-\mathcal{E}(\vec{f})|\nonumber\\
&\leq&\sup_{\vec{f}\in
\mathcal{F}_r}|\mathcal{E}_\mathcal{Z}(\vec{f})-\mathcal{E}_{\mathcal{Z}_i^\prime}(\vec{f})|\leq
\frac{4(2n-1)(M+\kappa Dr)^2}{n^2}.
\end{eqnarray}
Interchange the roles of $\mathcal{Z}$ and $\mathcal{Z}_i^\prime$
gives
$$|S(\mathcal{Z},r)-S(\mathcal{Z}_i^\prime,r)|\leq \frac{8(M+\kappa Dr)^2}{n}.$$
By McDiarmid's inequality, we obtain the desired estimate.
\end{proof}

In order to bound $S(\mathcal{Z},r)$ using Lemma \ref{lemma appendix four}, we need a bound of $ES(\mathcal{Z},r)$.
\begin{lemma}\label{lemma appendix three}
For every  $r>0$,
$$ES(\mathcal{Z},r)\leq \frac{11(\kappa Dr+M)^2}{\sqrt{n}}.$$
\end{lemma}
\begin{proof}
Denote
$\xi(\mathbf{x},y,\mathbf{u},v)=\omega^s(\mathbf{x}-\mathbf{u})(y-v+\vec{f}(\mathbf{x})(\mathbf{u}-\mathbf{x})).$
Then
$\mathcal{E}(\vec{f})=E_{(\mathbf{u},v)}E_{(\mathbf{x},y)}\xi(\mathbf{x},y,\mathbf{u},v)\}$
and
$\mathcal{E}_\mathcal{Z}(\vec{f})=\frac{1}{n^2}\sum_{i,j=1}^n\xi(\mathbf{x}_i,y_i,\mathbf{x}_j,y_j)
$. One can easily check that
\begin{eqnarray}
S(\mathcal{Z},r)&\leq& \sup_{\vec{f}\in
\mathcal{F}_r}|\mathcal{E}(\vec{f})-\frac{1}{n}\sum_{j=1}^nE_{(\mathbf{x},y)}\xi(\mathbf{x},y,\mathbf{x}_j,y_j)|
\nonumber\\&&+\sup_{\vec{f}\in
\mathcal{F}_r}\left|\frac{1}{n}\sum_{j=1}^nE_{(\mathbf{x},y)}\xi(\mathbf{x},y,\mathbf{x}_j,y_j)
-\mathcal{E}_\mathcal{Z}(\vec{f})\right|\nonumber\\&\leq&\sup_{\vec{f}\in
\mathcal{F}_r}E_{(\mathbf{x},y)}\left|E_{(\mathbf{u},v)}\xi(\mathbf{x},y,\mathbf{u},v)-\frac{1}{n}\sum_{j=1}^n
\xi(\mathbf{x},y,\mathbf{x}_j,y_j)\right|\nonumber\\&&+\frac{1}{n}\sum_{j=1}^n\sup_{\vec{f}\in\mathcal{F}_r}
\sup_{(\mathbf{u},v)\in
\mathcal{Z}}\left|E_{(\mathbf{x},y)}\xi(\mathbf{x},y,\mathbf{u},v)-\frac{1}{n-1}\sum_{i=1,i\neq
j}^n\xi(\mathbf{x}_i,y_i,\mathbf{u},v)\right|\nonumber\\&&+\frac{1}{n^2(n-1)}\sum_{j=1}^n\sum_{i\neq
j,i=1}^n\xi(\mathbf{x}_i,y_i,\mathbf{x}_j,y_j)\nonumber\\&:=&S_1+S_2+S_3.\nonumber
\end{eqnarray}
Let $\epsilon_i,i=1,\cdots,n$ be independent Rademacher variables.
For $S_1$, by using the properties of Rademacher complexities \citep{VW:book:1996}, we
have
\begin{eqnarray}
ES_1(\mathcal{Z})&=&E_{(\mathbf{x},y)}\sup_{\vec{f}\in
\mathcal{F}_r}\left|E_{(\mathbf{u},v)}\xi(\mathbf{x},y,\mathbf{u},v)-\frac{1}{n}\sum_{j=1}^n\xi(\mathbf{x},y,
\mathbf{x}_j,y_j)\right|\nonumber\\&\leq& 2\sup_{(\mathbf{x},y)\in
\mathcal{Z}}E\sup_{\vec{f}\in
\mathcal{F}_r}\left|\frac{1}{n}\sum_{j=1}^n\epsilon_j\omega^s(\mathbf{x}-\mathbf{x}_j)(y_j-y+\vec{f}(\mathbf{x}_j)(\mathbf{x}-\mathbf{x}_j))^2\right|\nonumber\\
&\leq& 4(M+\kappa Dr)\left(\sup_{(\mathbf{x},y)\in
\mathcal{Z}}E\sup_{\vec{f}\in
\mathcal{F}_r}\frac{1}{n}\sum_{j=1}^n\epsilon_j(y_j-y+\vec{f}(\mathbf{x}_j)(\mathbf{x}-\mathbf{x}_j))+\frac{M}{\sqrt{n}}\right)\nonumber\\&\leq&
\frac{5(\kappa Dr+M)^2}{\sqrt{n}}.\nonumber
\end{eqnarray}
Similarly, we can verify $ES_2(\mathcal{Z})\leq \frac{5(\kappa
Dr+M)^2}{\sqrt{n}}.$ Obviously, $S_3\leq \frac{(M+\kappa
Dr)^2}{n}.$ Combining the estimates for $S_1,S_2$ and $S_3$,
we can get the desired estimate $$ES(\mathcal{Z},r)\leq
\frac{10(\kappa Dr+M)^2}{\sqrt{n}}+\frac{(M+\kappa
Dr)^2}{n}\leq \frac{11(M+\kappa Dr)^2}{\sqrt{n}}.$$
\end{proof}
\begin{proposition}\label{prop appendix two}
Assume $r>1$. There exists a constant $C_3>0$ such that with
confidence at least $1-\delta$, $$\varphi(\mathcal{Z})\leq
C_3\frac{(\kappa Dr+M)^2\log\frac{2}{\delta}}{\sqrt{n}}.$$
\end{proposition}
\begin{proof}
The result is a direct application of inequality \eqref{inq sample
error bound one}, lemma \ref{lemma appendix four} and lemma
\ref{lemma appendix three}.
\end{proof}
Note that in order to use this Proposition, we still need a bound on $\Omega(\vec{f}_{\mathcal{Z}})=\lambda\sum_i\|f_\mathcal{Z}^i\|_\mathcal{K}$.
We first state a rough bound.
\begin{lemma}\label{lemma appendix five}
For every $s>0$ and $\lambda>0$, $\Omega(\vec{f}_\mathcal{Z})\leq
M^2.$\end{lemma}
\begin{proof}
The conclusion follows from the fact
$$\Omega(\vec{f}_{\mathcal{Z}})\leq \mathcal{E}_\mathcal{Z}(\vec{f}_{\mathcal{Z}})+ \Omega(\vec{f}_\mathcal{Z})\leq \mathcal{E}_\mathcal{Z}(\vec{0})
\leq M^2.$$
\end{proof}
However, using this quantity the bound in Theorem \ref{thm appendix one} is
at least of order  $O(\frac{1}{\lambda^2s^{2p+4-\theta}})$ which
tends to $\infty$ as $s\rightarrow 0$ and $\lambda\rightarrow 0.$ So
a sharper bound is needed.
It will be given in Section \ref{subsubsec convergence rate}.
\subsubsection{Approximation Error Estimation}
We now bound the approximation error $\mathcal{A}(\lambda).$
\begin{proposition}\label{prop appendix three}
If $\nabla f_\rho\in \mathbb{H}_\mathcal{K}^p,$ then
$\mathcal{A}(\lambda)\leq C_4(\lambda+s^{4+p})$ for some $C_4>0.$
\end{proposition}
\begin{proof}
By the definition of $\mathcal{A}(\lambda)$ and the fact that
$\nabla f_\rho\in \mathbb{H}_\mathcal{K}^p$,
$$\mathcal{A}(\lambda)\leq \mathcal{E}(\nabla f_\rho)-2\sigma_s^2+\Omega(\nabla f_\rho).$$
Since \begin{eqnarray} \mathcal{E}(\nabla f_\rho)-2\sigma_s^2
&=&\int_X\int_X\omega^s(\mathbf{x}-\mathbf{u})(f_\rho(\mathbf{x})-f_\rho(\mathbf{u})+\nabla
f_\rho(\mathbf{x})(\mathbf{u}-\mathbf{x}))^2d\rho_X(\mathbf{x})d\rho_X(\mathbf{u})\nonumber\\
&\leq& (C_\mathcal{K})^2c_\rho\int_X\int_X\omega^s(\mathbf{x}-\mathbf{u})\|\mathbf{u}-\mathbf{x}\|^4d\mathbf{u}d\rho_X(\mathbf{x})\nonumber\\
&\leq& (C_\mathcal{K})^2c_\rho M_4s^{4+p}.\nonumber
\end{eqnarray}
Taking $C_4=\max \{(C_\mathcal{K})^2c_\rho M_4, \sum_{i=1}^p\|(\nabla
f_\rho)^i\|_\mathcal{K}\},$ we get the desired  result.
\end{proof}

\subsubsection{Convergence rate}\label{subsubsec convergence rate}
Following directly from
Proposition \ref{prop appendix one}, Proposition
\ref{prop appendix two} and Proposition \ref{prop appendix three}, we get
 \begin{theorem} \label{thm appendix two}
If $\nabla f_\rho\in \mathbb{H}_\mathcal{K}^p, \vec{f}_\mathcal{Z}$
and $\vec{f}_\lambda$ are in $\mathcal{F}_r$ for some $r\geq 1$,
then with confidence $1-\delta$
$$\mathcal{E}(\vec{f}_\mathcal{Z})-2\sigma_s^2\leq C_2\left(\frac{(M+\kappa Dr)^2\log\frac{2}{\delta}}{\sqrt{n}}+s^{4+p}+\lambda\right),$$
where $C_2$ is a constant independent of $r, s$ or $\lambda$.
\end{theorem}

 In order to apply Theorem \ref{thm appendix one}, we need  a sharp bound
on $\Omega(\vec{f}_\mathcal{Z}):=\lambda\sum_i\|f^i_\mathcal{Z}\|_\mathcal{K}.$
 \begin{lemma}\label{lemma appendix seven}
Under the assumptions of Theorem \ref{thm error rates}, with
confidence at least $1-\delta$
$$\Omega(\vec{f}_{\mathcal{Z}})\leq C_5\left(\lambda+s^{4+p}+\left(1+\frac{\kappa DM}{\lambda
}\right)^2
\frac{M^2\log\frac{2}{\delta}}{\sqrt{n}}\right)$$ for
some $C_5>0$ independent of $s$ or $\lambda$.
\end{lemma}
\begin{proof}
By the fact $\mathcal{E}(\vec{f}_\mathcal{Z})-2\sigma_s^2>0$ and
Proposition \ref{prop appendix one}, we have
$\Omega(\vec{f}_\mathcal{Z})\leq
\frac{1}{\lambda}(\varphi(\mathcal{Z})+\mathcal{A}(\lambda)).$ Since
both $\vec{f}_\mathcal{Z}$ and $\vec{f}_\lambda$ are in
$\mathcal{F}_{\frac{M^2}{\lambda }}$,  using  Proposition
\ref{prop appendix two}, we have with probability at least
$1-\delta$,
$$\varphi(\mathcal{Z})\leq C_3\left(1+\frac{\kappa DM}{\lambda
}\right)^2
\frac{M^2\log\frac{2}{\delta}}{\sqrt{n}}.$$ Together
with Proposition \ref{prop appendix three}, we obtain the desired
estimate with $C_5=\max\{C_3,C_4\}.$
\end{proof}

 \begin{lemma}\label{lemma appendix six}
Under the assumptions of Theorem \ref{thm error rates},
$$\Omega(\vec{f}_\lambda)\leq C_4(\lambda+s^{4+p}),$$
where $C_4$ is a constant independent of $\lambda$ or $s$.
 \end{lemma}
 \begin{proof}
Since $\mathcal{E}(\vec{f}_\lambda)-2\sigma_s^2$ is non-negative for
all $\vec{f},$ we have $$\Omega(\vec{f}_\lambda)\leq
\mathcal{E}(\vec{f}_\lambda)-2\sigma_s^2+\lambda
\Omega(\vec{f}_\lambda)=\mathcal{A}(\lambda).$$ This in conjunction
with proposition \ref{prop appendix three} implies the conclusion.
 \end{proof}

Now we will use Theorem \ref{thm appendix one} and Theorem \ref{thm appendix
 two} to prove Theorem \ref{thm error rates}.

{\bf Proof of Theorem \ref{thm error rates}:} By Theorem \ref{thm
appendix one} and Theorem \ref{thm appendix two}, we have with at
least probability $1-\frac{\delta}{2},$
$$\|\vec{f}_\mathcal{Z}-\nabla f_\rho\|_{L^2_{\rho_X}}^2\leq C_0\left\{r^2s^\theta
+s^{2-\theta}+\frac{C_2}{s^{p+2+\theta}}\left(\frac{(M+\kappa
Dr)^2\log\frac{4}{\delta}}{\sqrt{n}}+s^{4+p}+\lambda\right)\right\},$$
if both $\vec{f}_\mathcal{Z}$ and $\vec{f}_\lambda$ are in
$\mathcal{F}_r$ for some $r>1$. By lemma \ref{lemma appendix six}
and lemma \ref{lemma appendix seven}, we can state that both
$\vec{f}_\mathcal{Z}$ and $\vec{f}_\lambda$ are in $\mathcal{F}_r$
with probability $1-\frac{\delta}{2}$ if $$
r=\max\left\{1+\frac{s^{4+p}}{\lambda},\left(1+\frac{\kappa DM}{\lambda
}\right)^2\frac{M^2\log\frac{4}{\delta}}{\lambda\sqrt{n}}\right\}.$$
Choose
$s=\left(\frac{1}{n}\right)^{\frac{1}{2(p+2+2\theta)}},\lambda=\left(\frac{1}{n}\right)^{\frac{\theta}{p+2+2\theta}},$
we obtain with confidence $1-\delta$,
$$\|\vec{f}_\mathcal{Z}-\nabla f_\rho\|_{L^2_{\rho_X}}\leq C\left(\frac{1}{n}\right)^{\frac{\theta}{4(p+2+2\theta)}}.$$

\subsection {Convergence Analysis in the Manifold Setting}
The convergence analysis in the Manifold Setting can be derived in a similar way as the one in the Euclidean setting.
The idea behind the proof  for the convergence of the
gradient consists of simultaneously controlling a sample or
estimation error term and a regularization or approximation error
term.

As done in the convergence analysis in the Euclidean setting, we first use the excess error, $\mathcal{E}(\vec{f})-2\sigma_s^2$, to
bound the $L_{\rho_X}^2$ differences of $\nabla_Xf_\rho$ and
$(d\Phi)^*(\vec{f})$.

Recall
$$\mathcal{F}_r=\{\vec{f}\in \mathcal{H}_\mathcal{K}^p:\sum_{i=1}^p\|f^i\|_\mathcal{K}\leq r\},\quad r>0.$$

\begin{theorem}\label{theorm comparative analysis manifold}
Assume $\rho_X$ satisfies the condition \eqref{inq marginal distribution} and
\eqref{inq density} and $\nabla_X f_\rho\in C^2(X)$. For $\vec{f}\in \mathcal{F}_r$ with some $r\geq 1$, there exist a constant $C_0>0$ such that
\begin{equation*}
\|(d\Phi)^*(\vec{f})-\nabla_Xf_\rho\|_{L_{\rho_X}^2}^2\leq C_0(r^2s^\theta+\frac{1}{s^{d+2+\theta}}(\mathcal{E}(\vec{f})-2\sigma_s^2)).
\end{equation*}
\end{theorem}
\begin{proof}
It can be directly derived from lemma B.1 in \cite{MWZ:Bernoulli:2009} by using the inequality $\sum_{i=1}^n |v_i|^2\leq (\sum_{i=1}^n |v_i|)^2$.
\end{proof}
\subsubsection{Excess Error Estimation}
In this subsection, we will bound $\mathcal{E}(\vec{f}_\mathcal{Z})-2\sigma_s^2$. First, we  decompose the excess error into sample error and
approximation error.
\begin{proposition}\label{prop appendix four}
Let $\vec{f}_\lambda$ be defined as \eqref{eq regularization function}, $$\varphi(\mathcal{Z})=(\mathcal{E}(\vec{f}_\mathcal{Z})-\mathcal{E}_\mathcal{Z}
(\vec{f}_\mathcal{Z}))+(\mathcal{E}_\mathcal{Z}(\vec{f}_\lambda)-\mathcal{E}
(\vec{f}_\lambda))$$ and
$$\mathcal{A}(\lambda)=\inf_{\vec{f}\in \mathbb{H}_\mathcal{K}^p}\left\{
\mathcal{E}(\vec{f})-2\sigma_s^2+\Omega(\vec{f})\right\}.$$ Then, we have
$$\mathcal{E}(\vec{f}_\mathcal{Z})-2\sigma_s^2+\Omega(\vec{f}_\mathcal{Z})\leq
\varphi(\mathcal{Z})+\mathcal{A}(\lambda).$$
\end{proposition}
Since the proof of Proposition \ref{prop appendix two} doesn't need any structure information of $X$,
it is still true in the manifold setting. Thus we have the same sample error bound as the one in the Euclidean setting.
What left is to give an estimate for the approximation error $\mathcal{A}(\lambda)$ in the manifold setting.
\begin{proposition}\label{prop appendix five} Let $X$ be a connected compact $C^\infty$ submanifold of $\mathbb{R}^p$ which is isometrically embedded and of dimension $d$. If $f_\rho\in C^2(X)$  and $d\Phi(\nabla_Xf_\rho)\in \mathcal{H}_\mathcal{K}^p$, then $$\mathcal{A}(\lambda)\leq C_6(\lambda+s^{4+d})$$ for some $C_6>0.$
\end{proposition}
\begin{proof}
By the definition of $\mathcal{A}(\lambda)$ and the fact that $d\Phi(\nabla_Xf_\rho)\in \mathcal{H}_\mathcal{K}^p$,
$$
\mathcal{A}(\lambda)\leq \mathcal{E}(d\Phi(\nabla_Xf_\rho))-2\sigma_s^2+\Omega(d\Phi(\nabla_Xf_\rho)).
$$
Note that $f_\rho\in C^2(X)$  and $d\Phi(\nabla_Xf_\rho)\in \mathcal{H}_\mathcal{K}^p$.
By Lemma B.2 in \cite{MWZ:Bernoulli:2009}, we have
$$
\mathcal{E}(d\Phi(\nabla_Xf_\rho))-2\sigma_s^2\leq C_7 s^{4+d},
$$
where $C_7$ is a constant independent of $s$.
Taking $C_6=\max\{C_7,\sum_{i=1}^p\|(d\Phi(\nabla_Xf_\rho))^i\|_\mathcal{K}\},$ we get the desired result.
\end{proof}
Combining Proposition \ref{prop appendix four}, Proposition \ref{prop appendix two} and Proposition \ref{prop appendix five}, we get the estimate for the excess error.
\begin{theorem}\label{theorem excess error manifold}
If $d\Phi(\nabla f_\rho)\in \mathbb{H}_\mathcal{K}^p, \vec{f}_\mathcal{Z}$ and $\vec{f}_\lambda$ are in $\mathcal{F}_r$ for some $r\geq 1$, then with confidence $1-\delta$,
\begin{equation*}
\mathcal{E}(\vec{f}_\mathcal{Z})-2\sigma_s^2\leq C_8\left(\frac{(M+\kappa Dr)^2\log\frac{2}{\delta}}{\sqrt{n}}+s^{d+4}+\lambda\right),
\end{equation*}
where $C_8$ is a constant independent of $s,\lambda,\delta$ or $r$.
\end{theorem}

\subsubsection{Convergence Rate}
In order to use Theorem \ref{theorm comparative analysis manifold} and Theorem \ref{theorem excess error manifold}, we need sharp  estimations for
$\sum_{i=1}^p\|(d\Phi(\nabla_Xf_\rho))^i\|_\mathcal{K}$ and $\sum_{i=1}^p\|f^i_\lambda\|_\mathcal{K}$. This can be done using the same argument as the one in the Euclidean setting, we omit the proof here.
\begin{lemma}\label{lemma bound manifold}
Under the assumptions of Theorem \ref{Theorem convergence rate manifold}, with confidence at least $1-\delta$,
\begin{equation*}
\Omega(\vec{f}_{\mathcal{Z}})\leq C_9\left(\lambda+s^{4+d}+\left(1+\frac{\kappa DM}{\lambda}\right)^2\frac{M^2\log\frac{2}{\delta}}{\sqrt{n}}\right)
\end{equation*} and
\begin{equation*}
  \Omega(\vec{f}_\lambda)\leq C_9(\lambda+s^{4+d}),
\end{equation*}
where $C_9$ is a constant independent of $\lambda$ or $s$.
\end{lemma}
Now we prove Theorem \ref{Theorem convergence rate manifold}.

{\bf Proof of Theorem \ref{Theorem convergence rate manifold}:} By the same argument as the one in proving Theorem \ref{thm error rates}, we can derive the convergence rate using Theorem \ref{theorm comparative analysis manifold}, Theorem \ref{theorem excess error manifold} and Lemma \ref{lemma bound manifold}.

\section{Algorithm for solving sparse gradient learning}\label{sec implementation}

In this section, we describe how to solve the optimization problem  in Eq.\ \eqref{gradient learning algorithm}.
Our overall strategy is to first transfer the convex functional from the infinite dimensional to a finite dimensional space by using the reproducing property of  RHKS, and then develop a forward-backward splitting algorithm to solve the reduced finite dimensional problem.

\subsection{From infinite dimensional to finite dimensional optimization}\label{sec discretization}
 Let $\mathcal{K}:\mathbb{R}^p\times
\mathbb{R}^p\rightarrow \mathbb{R}^p$ be continuous, symmetric and
positive semidefinite, i.e., for any finite set of distinct points
$\{\mathbf{x}_1,\cdots, \mathbf{x}_n\}\subset\mathbb{R}^p$, the
matrix
$\left[\mathcal{K}(\mathbf{x}_i,\mathbf{x}_j)\right]_{i,j=1}^n$ is
positive semidefinite \citep{Aron}. Such a function is called a {\it
Mercer kernel}. The RKHS $\mathbb{H}_{\mathcal{K}}$ associated with
the Mercer kernel $\mathcal{K}$ is defined  to be the completion of
the linear span  of the set of functions
$\{\mathcal{K}_\mathbf{x}:=\mathcal{K}(\mathbf{x},\cdot):\mathbf{x}\in
\mathbb{R}^n\}$ with the inner product $\langle
\cdot,\cdot\rangle_\mathcal{K}$ satisfying $\langle
\mathcal{K}_\mathbf{x},\mathcal{K}_\mathbf{u}\rangle_\mathcal{K}=\mathcal{K}(\mathbf{x},\mathbf{u}).$
The reproducing property of $\mathbb{H}_{\mathcal{K}}$ states that
\begin{equation}
\langle
\mathcal{K}_\mathbf{x},h\rangle_\mathcal{K}=h(\mathbf{x})\qquad\forall
\mathbf{x}\in \mathbb{R}^p,h\in
\mathbb{H}_\mathcal{K}.\label{reproducing property}
\end{equation}

By the reproducing property \eqref{reproducing property}, we have
the following representer theorem, which states that the solution of
\eqref{gradient learning algorithm} exists and  lies in the finite
dimensional space spanned by
$\{\mathcal{K}_{\mathbf{x}_i}\}_{i=1}^n$.  Hence the sparse gradient learning in Eq.\ \eqref{gradient learning algorithm} can be converted into a finite dimensional optimization problem. The proof of the theorem  is
standard and  follows the same line as done in \citep{SS:book:2002,MZH:JMLR:2006}.

\begin{theorem}\label{theorem representer}
Given a data set $\mathcal{Z}$, the solution of Eq.\ \eqref{gradient
learning algorithm} exists and takes the following form
\begin{equation}
f^j_{\mathcal{Z}}(\mathbf{x})=\sum_{i=1}^n
c^j_{i,\mathcal{Z}}\mathcal{K}(\mathbf{x},\mathbf{x}_i),\label{representer
equation}
\end{equation}
where $c^j_{i,\mathcal{Z}}\in\mathbb{R}$ for $j=1,\ldots,p$ and
$i=1,\ldots,n$.
\end{theorem}

\begin{proof}
The existence follows from the convexity of  functionals ${\cal E}_{\mathcal{Z}}(\vec{f})$ and $\Omega(\vec{f})$.
Suppose $\vec{f}_{\mathcal{Z}}$ is a minimizer. We can
write functions $\vec{f}_{\mathcal{Z}}\in {\cal H}_K^p$ as
$$\vec{f}_{\mathcal{Z}}=\vec{f}_{\|}+\vec{f}_{\bot},$$
where each element of $\vec{f}_{\|}$ is in the span of
$\{K_{\mathbf{x}_1},\cdots,K_{\mathbf{x}_n}\}$ and $\vec{f}_{\bot}$ are functions in
the orthogonal complement. The reproducing property yields
$\vec{f}(\mathbf{x}_i)=\vec{f}_{\|}(\mathbf{x}_i)$ for all $\mathbf{x}_i$. So the functions
$\vec{f}_{\bot}$ do not have an effect on ${\cal E}_{\mathcal{Z}}(\vec{f})$. But $\|\vec{f}_{\mathcal{Z}}\|_K=
\|\vec{f}_{\|}+\vec{f}_{\bot}\|_K>\|\vec{f}_{\|}\|_K$ unless
$\vec{f}_{\bot}=0$. This implies that $\vec{f}_{\mathcal{Z}}=\vec{f}_{\|}$, which leads to the representation of
$\vec{f}_{\mathcal{Z}}$ in Eq.\ \eqref{representer equation}.
\end{proof}

Using Theorem \ref{theorem representer}, we can transfer the
infinite dimensional minimization problem \eqref{gradient learning
algorithm} to an finite dimensional one. Define the matrix
$C_{\mathcal{Z}}:=[c^j_{i,\mathcal{Z}}]_{j=1,i=1}^{p,n}\in\mathbb{R}^{p\times
n}$. Therefore, the optimization problem in \eqref{gradient learning
algorithm} has only $p\times n$ degrees of freedom, and is actually
an optimization problem in terms of a coefficient matrix
$C:=[c_{i}^{j}]_{j=1,i=1}^{p,n}\in\mathbb{R}^{p\times n}$. Write $C$
into column vectors as $C:=(\mathbf{c}_1,\ldots,\mathbf{c}_n)$ with
$\mathbf{c}_i\in\mathbb{R}^{p}$ for $i=1,\cdots,n$, and into row
vectors as $C:=(\mathbf{c}^1,\ldots,\mathbf{c}^p)^T$ with
$\mathbf{c}^j\in\mathbb{R}^{n}$ for $j=1,\cdots,p$. Let the kernel
matrix be
$K:=[\mathcal{K}(\mathbf{x}_i,\mathbf{x}_j)]_{i=1,j=1}^{n,n}\in\mathbb{R}^{n\times
n}$. After expanding each component $f^j$ of $\vec{f}$ in \eqref{gradient
learning algorithm} as
$f^j(\mathbf{x})=\sum_{i=1}^nc^j_i\mathcal{K}(\mathbf{x},\mathbf{x}_i)$, the objective function in Eq.\ \eqref{gradient learning algorithm} becomes a function of $C$ as
\begin{eqnarray}\label{discrete energy}
\Phi(C)&=&{\cal E}_{\mathcal{Z}}(\vec{f})+\Omega(\vec{f})\cr
&=&\frac{1}{n^2}\sum_{i,j=1}^n
\omega_{i,j}^s\big(y_i-y_j+\sum_{k=1}^p\sum_{\ell=1}^nc_\ell^k\mathcal{K}(\mathbf{x}_i,\mathbf{x}_\ell)(x_j^k-x_i^k)\big)^2
+\lambda\sum_{j=1}^p\sqrt{\sum_{i,k=1}^nc_{i}^{j}\mathcal{K}(\mathbf{x}_i,\mathbf{x}_k)c_{k}^{j}}\cr
&=&\frac{1}{n^2}\sum_{i,j=1}^n
\omega_{i,j}^s\big(y_i-y_j+\sum_{\ell=1}^n\mathcal{K}(\mathbf{x}_\ell,\mathbf{x}_i)(\mathbf{x}_j-\mathbf{x}_i)^T\mathbf{c}_\ell\big)^2
+\lambda\sum_{j=1}^p\sqrt{(\mathbf{c}^j)^TK\mathbf{c}^j}\cr
&=&\frac{1}{n^2}\sum_{i,j=1}^n
\omega_{i,j}^s\big(y_i-y_j+(\mathbf{x}_j-\mathbf{x}_i)^TC\mathbf{k}_i\big)^2
+\lambda\sum_{j=1}^p\sqrt{(\mathbf{c}^j)^TK\mathbf{c}^j},
\end{eqnarray}
where $\mathbf{k}_i\in\mathbb{R}^n$ is the $i$-th column of $K$,
i.e., $K=(\mathbf{k}_1,\ldots,\mathbf{k}_n)$. Then, by Theorem
\ref{theorem representer},
\begin{equation}\label{min Phi}
C_{\mathcal{Z}}=\arg\min_{C\in\mathbb{R}^{p\times n}}\Phi(C).
\end{equation}

\subsection{Change of optimization variables}
The objective function $\Phi(C)$ in the reduced finite dimensional problem convex is a non-smooth function.
As such, most of the standard convex optimization techniques, such as gradient descent, Newton's method, etc,
cannot be directly applied. We will instead develop a forward-backward splitting algorithm to solve the problem.
For this purpose, we fist convert the problem into a simpler form by changing the optimization variables.

Note that $K$ is symmetric and positive semidefinite, so its square root $K^{1/2}$ is also symmetric
and positive semidefinite, and can be easily calculated. Denote the $i$-th column of $K^{1/2}$ by
$\mathbf{k}_i^{1/2}$, i.e.,
$K^{1/2}=(\mathbf{k}_1^{1/2},\ldots,\mathbf{k}_n^{1/2})$. Let
$\widetilde{C}=CK^{1/2}$ and write
$\widetilde{C}=(\widetilde{\mathbf{c}}_1,\ldots,\widetilde{\mathbf{c}}_n)=(\widetilde{\mathbf{c}}^1,\ldots,\widetilde{\mathbf{c}}^p)^T$,
where $\widetilde{\mathbf{c}}_i$ and $\widetilde{\mathbf{c}}^j$ are
the $i$-th column vector and $j$-th row vector respectively. Then
$\Phi(C)$ in Eq.\ \eqref{discrete energy} can be rewritten as a function of
$\widetilde{C}$
\begin{equation}\label{Psi}
\Psi(\widetilde{C})=\frac{1}{n^2}\sum_{i,j=1}^n
\omega_{i,j}^s\big(y_i-y_j+(\mathbf{x}_j-\mathbf{x}_i)^T\widetilde{C}\mathbf{k}_i^{1/2}\big)^2+\lambda\sum_{j=1}^p\|\widetilde{\mathbf{c}}^j\|_{2},
\end{equation}
where $\|\cdot\|_2$ is the Euclidean norm of $\mathbb{R}^{p}$.
Thus finding a solution $C_{\mathcal{Z}}$ of \eqref{min Phi}
is equivalent to identifying
\begin{equation}\label{min Psi}
\widetilde{C}_{\mathcal{Z}}=\arg\min_{\widetilde{C}\in\mathbb{R}^{p\times
n}}\Psi(\widetilde{C}),
\end{equation}
followed by setting
$C_{\mathcal{Z}}=\widetilde{C}_{\mathcal{Z}}K^{-1/2}$, where
$K^{-1/2}$ is the (pseudo) inverse of $K^{1/2}$ when $K$ is (not)
invertible.

Given matrix $\widetilde{C}_{\mathcal{Z}}$, the variables selected by the sparse gradient learning as defined in
Eq.\ \eqref{index set} is simply
\begin{equation}
S=\{j: \| \widetilde{\mathbf{c}}^j\|_2 \ne 0, j=1,\cdots,n\}.
\end{equation}
And similarly, the S-EDR directions can also be
directly derived from $\widetilde{C}_{\mathcal{Z}}$ by noting that the sparse gradient covariance matrix is equal to
\begin{equation}
\Xi=C^T_{\mathcal{Z}}KC_\mathcal{Z}=\widetilde{C}_\mathcal{Z}^T\widetilde{C}_\mathcal{Z}.
\end{equation}

\subsection{Forward-backward splitting algorithm}\label{sec algorithms}

Next we propose a forward-backward splitting to solve Eq.\ \eqref{min Psi}.  The forward-backward splitting is
commonly used to solve the $\ell_1$ related optimization problems in
machine learning \citep{LLZhang:NIPS:2008} and image processing
\citep{DDD:CPAM:04,CCS:ACHA:2008}. Our algorithm is derived from the general formulation described in \citep{CW:MMS:05}.

We first split the objective function $\Psi$ into a smooth term and a non-smooth term. Let
$\Psi=\Psi_1+\Psi_2$, where
$$
\Psi_1(\widetilde{C})=\lambda\sum_{i=1}^p\|\widetilde{\mathbf{c}}^i\|_{2}\qquad\mbox{and}\qquad
\Psi_2(\widetilde{C})=\frac{1}{n^2}\sum_{i,j=1}^n
\omega_{i,j}^s\big(y_i-y_j+(\mathbf{x}_j-\mathbf{x}_i)^T\widetilde{C}\mathbf{k}_i^{1/2}\big)^2
.
$$
The forward-backward splitting algorithm works by iteratively updating $\widetilde{C}$. Given a current estimate
$\widetilde{C}^{(k)}$, the next one is updated according to
\begin{equation}\label{PFBS for Psi}
\widetilde{C}^{(k+1)}=\mathrm{prox}_{\delta\Psi_1}(\widetilde{C}^{(k)}-\delta\nabla\Psi_2(\widetilde{C}^{(k)})),
\end{equation}
where $\delta>0$ is the step size, and $\mathrm{prox}_{\delta\Psi_1}$ is {\em a proximity operator} defined by
\begin{equation}\label{prox}
\mathrm{prox}_{\delta\Psi_1}(D)=\arg\min_{\widetilde{C}\in\mathbb{R}^{p\times
n}} \frac12\|D-\widetilde{C}\|_F^2+\delta\Psi_1(\widetilde{C}),
\end{equation}
where $\|\cdot\|_F$ is the Frobenius norm of $\mathbb{R}^{p\times
n}$.

To implement the algorithm \eqref{PFBS for Psi}, we need to know both $\nabla\Psi_2$ and
$\mathrm{prox}_{\delta\Psi_1}(\cdot)$. The term $\nabla\Psi_2$ is relatively easy to obtain,
\begin{equation}\label{nabla Psi2}
\nabla\Psi_2(\widetilde{C})
=\frac{2}{n^2}\sum_{i,j=1}^n\omega_{i,j}^s\big(y_i-y_j+(\mathbf{x}_j-\mathbf{x}_i)^T
\widetilde{C}\mathbf{k}_i^{1/2}\big)(\mathbf{x}_j-\mathbf{x}_i)(\mathbf{k}_i^{1/2})^T.
\end{equation}
The proximity operator $\mathrm{prox}_{\delta\Psi_1}$ is given in
the following lemma.

\begin{lemma}\label{lemma proximity operator}
Let $T_{\lambda\delta}(D)=\mathrm{prox}_{\delta\Psi_1}(D)$, where
$D=(\mathbf{d}^1,\ldots,\mathbf{d}^p)^T$ with $\mathbf{d}^j$ being
the $j$-th row vector of $D$. Then
\begin{equation}\label{thresh TD}
T_{\lambda\delta}(D)=\big(t_{\lambda\delta}(\mathbf{d}^1),\ldots,t_{\lambda\delta}(\mathbf{d}^p)\big)^T,
\end{equation}
where
\begin{equation}\label{thresh td}
t_{\lambda\delta}(\mathbf{d}^j)=\begin{cases}\mathbf{0},&\mbox{if}\quad\|\mathbf{d}^j\|_2\leq\lambda\delta,\cr
\frac{\|\mathbf{d}^j\|_2-\lambda\delta}{\|\mathbf{d}^j\|_2}\mathbf{d}^j,&\mbox{if}\quad\|\mathbf{d}^j\|_2>\lambda\delta.
\end{cases}
\end{equation}
\end{lemma}
\begin{proof}
From
\eqref{prox}, one can easily see that the row vectors
$\widetilde{\mathbf{c}}^j$, $j=1,\ldots,n$, of $\widetilde{C}$ are
independent of each others. Therefore, we have
\begin{equation}\label{prox 1D}
t_{\lambda\delta}(\mathbf{d}^j)=\arg\min_{\mathbf{c}\in\mathbb{R}^{n}}\frac12\|\mathbf{d}^j-\mathbf{c}\|_2^2+\lambda\delta\|\mathbf{c}\|_2.
\end{equation}
The energy function in the above minimization problem is strongly
convex, hence has a unique minimizer. Therefore, by the
subdifferential calculus (c.f. \citep{HL:BOOK:93}),
$t_{\lambda\delta}(\mathbf{d}^j)$ is the unique solution of the
following equation with unknown $\mathbf{c}$
\begin{equation}\label{first-order optimality}
\mathbf{0}\in\mathbf{c}-\mathbf{d}^j+\lambda\delta\partial(\|\mathbf{c}\|_2),
\end{equation}
where
$$
\partial(\|\mathbf{c}\|_2)=\{\mathbf{p}:\mathbf{p}\in\mathbb{R}^{n};~\|\mathbf{u}\|_2-\|\mathbf{c}\|_2
-(\mathbf{u}-\mathbf{c})^T\mathbf{p}\geq 0,~\forall
\mathbf{u}\in\mathbb{R}^{n}\}
$$
is the subdifferential of the function $\|\mathbf{c}\|_2$. If
$\|\mathbf{c}\|_2>0$, the function $\|\mathbf{c}\|_2$ is
differentiable, and its subdifferential contains only its gradient,
i.e.,
$\partial(\|\mathbf{c}\|_2)=\{\frac{\mathbf{c}}{\|\mathbf{c}\|_2}\}$.
If $\|\mathbf{c}\|_2=0$, then
$\partial(\|\mathbf{c}\|_2)=\{\mathbf{p}:
\mathbf{p}\in\mathbb{R}^{n};~\|\mathbf{u}\|_2-\mathbf{u}^T\mathbf{p}\geq
0,~\forall \mathbf{u}\in\mathbb{R}^{n}\}$. One can check that
$\partial(\|\mathbf{c}\|_2)=\{\mathbf{p}:
\mathbf{p}\in\mathbb{R}^{n}; \|\mathbf{p}\|_2\leq 1\}$ for this
case. Indeed, for any  vector $\mathbf{p}\in\mathbb{R}^n$ with
$\|\mathbf{p}\|_2\leq 1$,
$\|\mathbf{u}\|_2-\mathbf{u}^T\mathbf{p}\geq 0$ by the
Cauchy-Schwartz inequality. On the other hand, if there is an
element $\mathbf{p}$ of $\partial(\|\mathbf{c}\|_2)$ such that
$\|\mathbf{p}\|_2>1$, then, by setting $\mathbf{u}=\mathbf{p}$, we
get
$\|\mathbf{p}\|_2-\mathbf{p}^T\mathbf{p}=\|\mathbf{p}\|_2(1-\|\mathbf{p}\|_2)<0$,
which contradicts the definition of $\partial(\|\mathbf{c}\|_2)$. In
summary,
\begin{equation}\label{subdiff cnorm}
\partial(\|\mathbf{c}\|_2)=\begin{cases}
\{\frac{\mathbf{c}}{\|\mathbf{c}\|_2}\},&\mbox{if}\quad
\|\mathbf{c}\|_2>0,\cr \{\mathbf{p}: \mathbf{p}\in\mathbb{R}^{n};
\|\mathbf{p}\|_2\leq 1\},&\mbox{if}\quad \|\mathbf{c}\|_2=0.\cr
\end{cases}
\end{equation}
With \eqref{subdiff cnorm}, we see that
$t_{\lambda\delta}(\mathbf{d}^j)$ in \eqref{thresh td} is a solution
of \eqref{first-order optimality} hence \eqref{thresh TD} is
verified.
\end{proof}

Now, we obtain the following forward-backward splitting algorithm to find the optimal $\widetilde{C}$
in Eq. \eqref{min Phi}. After choosing a random initialization, we update $\widetilde{C}$ iteratively until convergence according to
\begin{equation}\label{iteration minimization}
\begin{cases}
D^{(k+1)}=\widetilde{C}^{(k)}-
\frac{2\delta}{n^2}\sum_{i,j=1}^n\omega_{i,j}^s\big(y_i-y_j+(\mathbf{x}_j-\mathbf{x}_i)^T\widetilde{C}^{(k)}\mathbf{k}_i^{1/2}\big)
(\mathbf{x}_j-\mathbf{x}_i)(\mathbf{k}_i^{1/2})^T,\cr
\widetilde{C}^{(k+1)}=T_{\lambda\delta}(D^{(k+1)}).
\end{cases}
\end{equation}

The iteration alternates between two steps:  1) an empirical error minimization step, which minimizes the empirical error
 $\mathcal{E}_{\mathcal{Z}}(\vec{f})$ along gradient descent directions; and 2) a variable selection step, implemented by the proximity operator $T_{\lambda\delta}$ defined in
\eqref{thresh TD}.  If the norm of the $j$-th row of $D^{(k)}$, or
correspondingly the norm $\|f^j\|_\mathcal{K}$ of the $j$-th partial
derivative, is smaller than a threshold $\lambda\delta$, the $j$-th row of $D^{(k)}$ will be set
 to $0$, i.e., the $j$-th variable is not selected.  Otherwise, the $j$-th row of $D^{(k)}$ will be kept unchanged except to reduce its norm by the threshold $\lambda\delta$.

Since $\Psi_2(\widetilde{C})$ is a quadratic
function of the entries of $\widetilde{C}$, the operator
norm of its Hessian $\|\nabla^2\Psi_2\|$ is a constant. Furthermore,
since the function $\Psi_2$ is coercive, i.e.,
$\|\widetilde{C}\|_F\to\infty$ implies that
$\Psi(\widetilde{C})\to\infty$, there exists at least one solution
of \eqref{min Psi}. By applying the convergence theory for the forward-backward splitting algorithm in \citep{CW:MMS:05}, we obtain the
following theorem.

\begin{theorem}\label{theorem convergence of PFBS}
If  $0<\delta<\frac{2}{\|\nabla^2\Psi_2\|}$, then the iteration \eqref{iteration
minimization} is guaranteed to converge to a solution of  Eq.\ \eqref{min Psi} for any
initialization $\widetilde{C}^{(0)}$.
\end{theorem}

The regularization parameter $\lambda$ controls the sparsity of the optimal solution. When $\lambda=0$, no sparsity constraint is imposed, and all variables will be selected. On the other extreme, when $\lambda$ is sufficiently large, the optimal solution will be $\tilde C=0$, and correspondingly none of the variables will be selected.  The following theorem  provides an upper bound of $\lambda$ above which no variables will be selected.  In practice, we choose $\lambda$ to be a number between $0$ and the upper bound usually through cross-validation.

\begin{theorem}\label{theorem regularization parameter}
Consider the sparse gradient learning in Eq.\ \eqref{min Psi}. Let
\begin{equation}
\lambda_{max} = \max_{1\le k\le p} \frac{2}{n^2} \left\|\sum_{i,j=1}^n\omega_{i,j}^s(y_i-y_j)(x_i^k-x_j^k)\mathbf{k}_i^{1/2}  \right \|_2
\end{equation}
Then the optimal solution is $\tilde C=0$ for all $\lambda \ge \lambda_{max}$, that is, none of the variables will be selected.
\end{theorem}
\begin{proof}
Obviously, if $\lambda=\infty$, the minimizer of Eq. \eqref{Psi}
is a $p\times n$ zero matrix.

When $\lambda<\infty$, the minimizer of Eq. \eqref{Psi} could also be a $p\times n$ zero matrix as long as $\lambda$ is large enough. Actually,
from iteration \eqref{iteration minimization}, if we choose $C^{(0)}=0$, then
$$D^{(1)}=-\frac{2\delta}{n^2}\sum_{i,j=1}^n\omega_{i,j}^s(y_i-y_j)(\mathbf{x}_j-\mathbf{x}_i)(\mathbf{k}_i^{\frac{1}{2}})^T$$
and $\widetilde{C}^{(1)}=T_{\lambda\delta}(D^{(1)}).$

Let $$\lambda_{max}=\max_{1\leq k\leq p}\frac{2}{n^2}\left\|\sum_{i,j=1}^n\omega_{i,j}^s(y_i-y_j)(\mathbf{x}_j^k-\mathbf{x}_i^k)(\mathbf{k}_i^{\frac{1}{2}})^T\right\|_2.$$
Then for any $\lambda\geq \lambda_{max}$, we have $\widetilde{C}^{(1)}=\mathbf{0}_{p\times n}$ by the definition of $T_{\lambda\delta}$. By induction,
$\widetilde{C}^{(k)}=\mathbf{0}_{p\times n}$ and the algorithm converge to $\widetilde{C}^{(\infty)}=\mathbf{0}_{p\times n}$ which is a minimizer of Eq. $\eqref{Psi}$ when $0<\delta<\frac{2}{\|\nabla^2 \Psi_2\|}$.
We get the desired result.
\end{proof}
\begin{remark}
In the proof of Theorem \ref{theorem regularization parameter}, we choose $C^{(0)}=\mathbf{0}_{p\times n}$ as the initial value of  iteration \eqref{iteration minimization} for simplicity. Actually, our argument is true for any initial value as long as $0<\delta<\frac{2}{\|\nabla^2 \Psi_2\|}$ since the algorithm converges to the minimizer of Eq. \eqref{Psi} when $0<\delta<\frac{2}{\|\nabla^2 \Psi_2\|}$.  Note that the convergence is independent of the choice of the initial value.
\end{remark}

It is not the first time to combine an iterative algorithm with a thresholding step to derive solutions with sparsity (see, e.g., \citep{DDD:CPAM:04}). However, different from the previous work, the sparsity we focus here is a block sparsity, that is, the row vectors of
$C$ (corresponding to partial derivatives $f^j$) are zero or nonzero vector-wise.
As such, the thresholding step in \eqref{thresh TD} is performed
row-vector-wise, not entry-wise as in the usual soft-thresholding
operator \citep{Don:IEEE:95}.

\subsection{Matrix size reduction}

The iteration in Eq.\ \eqref{iteration minimization} involves a weighted summation of $n^2$ number of $p \times n$ matrices
as defined by $(\mathbf{x}_j-\mathbf{x}_i)(\mathbf{k}_i^{1/2})^T$. When the dimension of the data is large, these matrices
are big, and could greatly influence the efficiency of the algorithm. However, if the number of samples is small,
that is, when $n<<p$, we can improve the efficiency of the algorithm by introducing a transformation to reduce the size of these matrices.

The main motivation is to note that the matrix
\begin{equation*}
\mathbf{M}_{\mathbf{x}}:=(\mathbf{x}_1-\mathbf{x}_n,\mathbf{x}_2-\mathbf{x}_n,\ldots,\mathbf{x}_{n-1}-\mathbf{x}_n,
\mathbf{x}_n-\mathbf{x}_n)\in\mathbb{R}^{p\times n}
\end{equation*}
is of low rank when $n$ is small. Suppose the rank of $\mathbf{M}_{\mathbf{x}}$ is $t$, which is no higher than $\min(n-1,p)$.

We use  singular value
decomposition to matrix $\mathbf{M}_{\mathbf{x}}$ with economy size. That is, $\mathbf{M}_{\mathbf{x}}
=U\Sigma V^T$, where $U$ is a $p \times n$
unitary matrix, $V$ is $n \times n$ unitary matrix, and $\Sigma=\hbox{diag}(\sigma_1,\ldots,\sigma_t,0,\ldots,0)\in \mathbb{R}^{n\times n}.$
Let $\beta=\Sigma V^T$, then
\begin{equation}\label{eq Sigmax decomposition}
\mathbf{M}_{\mathbf{x}}=U\beta.
\end{equation}

Denote $\beta=(\beta_1,\ldots,\beta_n)$. Then $\mathbf{x}_j-\mathbf{x}_i=U(\beta_j-\beta_i).$ Using these notations, the equation \eqref{iteration minimization}
is equivalent to
\begin{equation}\label{iteration minimization two}
\begin{cases}
D^{(k+1)}=\widetilde{C}^{(k)}-
\frac{2\delta U}{n^2}\sum_{i,j=1}^n\omega_{i,j}^s\big(y_i-y_j+(\mathbf{x}_j-\mathbf{x}_i)^T\widetilde{C}^{(k)}\mathbf{k}_i^{1/2}\big) (\beta_j-\beta_i)(\mathbf{k}_i^{1/2})^T,\cr
\widetilde{C}^{(k+1)}=T_{\lambda\delta}(D^{(k+1)}).
\end{cases}
\end{equation}
Note that now the second term in the right hand side of the first equation in
 \eqref{iteration minimization two} involves the summation of  $n^2$ number of
  $n\times n$ matrix rather than $p\times n$ matrices. Furthermore, we calculate the first iteration of Eq.
\eqref{iteration minimization two} using two steps: 1) we calculate  $y_i-y_j+(\mathbf{x}_j-\mathbf{x}_i)^T
\widetilde{C}^{(k)}\mathbf{k}_i^{1/2}$ and store it in an $n\times n$ matrix $r$; 2) we calculate the first
iteration of Eq.  \eqref{iteration minimization two} using the value $r(i,j)$. These two strategies greatly improve the
efficiency of the algorithm when $p>>n$. More specifically, we reduce the update for $D^{(k)}$ in Eq. \eqref{iteration minimization} of complexity $O(n^3p)$ into a problem of complexity $O(n^2p+n^4)$. A detailed implementation of the algorithm is shown in Algorithm  1.

\begin{remark}
Each update in Eq. \eqref{iteration minimization} involves the summation of $n^2$ terms, which could be inefficient for datasets with large number of samples. A strategy to reduce the number of computations is to use a truncated weight function, e.g.,
\begin{equation}\label{eq:truncated-weight}
\omega_{ij}^s=\left\{\begin{array}{ll}\exp(-\frac{2\|\mathbf{x}_i-\mathbf{x}_j\|^2}{s^2}),&\mathbf{x}_j\in\mathcal{N}_i^k,\\
0,&\hbox{otherwise},\end{array}\right.
\end{equation}
where  $\mathcal{N}_i^k=\{\mathbf{x}_j:\mathbf{x}_j \ \hbox{is in the $k$ nearest neighborhood of} \ \mathbf{x}_i\}$. This can reduce the number of summations from $n^2$ to $kn$.
\end{remark}

\begin{algorithm}[htb]\label{table algorithm}
\caption{Forward-backward splitting algorithm to solve sparse
gradient learning for regression.}{}\centering\fbox{
\begin{minipage}{.9\textwidth}
\alginout{data \{$\mathbf{x}_i,y_i\}_{i=1}^n$, kernel
$\mathcal{K}(\mathbf{x},\mathbf{y})$, weight function $\omega^s(\mathbf{x},\mathbf{y})$, parameters
$\delta, \lambda$ and matrix $\widetilde{C}^{(0)}$.}{the selected
variables $\mathcal{S}$ and S-EDRs.}
\begin{enumerate}
\item Compute $K$, $K^{1/2}$. Do the singular value decomposition with economy size for the matrix
$\mathbf{M}_{\mathbf{x}}=(\mathbf{x}_1-\mathbf{x}_n,\ldots,\mathbf{x}_n-\mathbf{x}_n)$ and get
$\mathbf{M}_{\mathbf{x}}=U\Sigma V^T$. Denote $\beta=(\beta_1,\ldots,\beta_n)=\Sigma V^T.$ Compute
$G_{ij}=\omega_{i,j}^s(\beta_j-\beta_i)(\mathbf{k}_i^{1/2})^T,
i=1,\ldots,n,j=1,\ldots,n$ and let $k=0$.
\item {\bf While} the convergence condition is not true {\bf do}
\begin{enumerate}
\item Compute the residual
$r^{(k)}=(r^{(k)}_{ij})\in\mathbb{R}^{n\times n}$, where
$r^{(k)}_{ij}=y_i-y_j+(\mathbf{x}_j-\mathbf{x}_i)^T\widetilde{C}^{(k)}\mathbf{k}_i^{1/2}$.
\item Compute
$g^{(k)}=\frac{2}{n^2}\sum_{i,j=1}^{n}r^{(k)}_{ij}G_{ij}$.
\item Set $D^{(k)}=\widetilde{C}^{(k)}-\delta U g^{(k)}$.
For the row vectors $(\mathbf{d}^{i})^{(k)}$, $i=1,\ldots,p$, of
$D^{(k)}$, perform the variable selection procedure
according to \eqref{thresh td} to get row vectors
$(\widetilde{\mathbf{c}}^i)^{(k+1)}$ of $\widetilde{C}^{(k+1)}$.
\begin{enumerate}
\item If $\|(\mathbf{d}^{i})^{(k)}\|_2\leq\lambda\delta$, the variable is
not selected, and we set $(\widetilde{\mathbf{c}}^i)^{(k+1)}=0$.
\item If $\|(\mathbf{d}^{i})^{(k)}\|_2>\lambda\delta$, the
variable is selected, and we set
$$(\widetilde{\mathbf{d}}^i)^{(k+1)}=\frac{\|(\mathbf{d}^{i})^{(k)}\|_2-\lambda\delta}{\|(\mathbf{d}^{i})^{(k)}\|_2}(\mathbf{d}^{i})^{(k)}.$$
\end{enumerate}
\item Update
$\widetilde{C}^{(k+1)}=\big((\widetilde{\mathbf{c}}^1)^{(k+1)},\ldots,(\widetilde{\mathbf{c}}^n)^{(k+1)}\big)^T$,
and set $k=k+1$.
\end{enumerate}

{\bf end while}
\item Variable selection: $\mathcal{S}=\{i: (\widetilde{\mathbf{c}}^i)^{(k+1)}\neq 0\}.$
\item Feature extraction:
let S-EGCM $\Xi=\widetilde{C}^{(k+1)}\cdot(\widetilde{C}^{(k+1)})^T$
and compute its eigenvectors via singular value decomposition of $\widetilde{C}^{(k+1)}$, we get the desired S-EDRs.
\end{enumerate}
\end{minipage}}
\end{algorithm}

\section{Sparse gradient learning for classification}\label{sec sparse gradient for classification}
In this section, we extend the sparse gradient learning algorithm from regression to classification problems. We will also
briefly introduce an implementation.

\subsection{Defining objective function}

Let $\mathbf{x}$  and $y\in\{-1,1\}$ be respectively  $\mathbb{R}^p$-valued  and binary random variables. The problem of classification is to estimate a classification function $f_C(\mathbf{x})$ from a set of observations  $\mathcal{Z}:=\{(\mathbf{x}_i,y_i)\}_{i=1}^n$,
where $\mathbf{x}_i:=(x_i^1,\ldots,x_i^p)^T\in \mathbb{R}^p$ is an
input, and $y_i\in\{-1,1\}$ is the corresponding output. A real valued
function $f_\rho^\phi :X\mapsto \mathbb{R}$ can be used to generate a classifier $f_C(\mathbf{x})=sgn(f_\rho^\phi(\mathbf{x})),$ where
$$sgn(f_\rho^\phi(\mathbf{x}))=\left\{\begin{array}{ll}1,&\hbox{if}\ f_\rho^\phi(\mathbf{x})>0,\\0,&\hbox{otherwise}.\end{array}\right.$$

Similar to regression, we also define an objective function, including a data fitting term and a regularization term, to learn the gradient of  $f_\rho^\phi$.  For classical  binary classification, we commonly use a
convex loss function $\phi(t)=\log(1+e^{-t})$ to learn $f^\phi_\rho$ and define the data fitting term to be $\frac{1}{n}\sum_{i=1}^n\phi(y_i f_\rho^\phi(\mathbf{x}_i))$.  The usage of loss function $\phi(t)$ is mainly motivated by
the fact that the optimal $f_\rho^\phi(\mathbf{x})=\log[P(y=1|\mathbf{x})/P(y=-1|\mathbf{x})]$, representing the log odds ratio between the two posterior probabilities. Note that the gradient of $f_\rho^\phi$ exists under very mild conditions.

As in the case of regression, we use the first order Taylor expansion to approximate the classification function $f_\rho^\phi$ by
$f_\rho^\phi(\mathbf{x})\approx f_\rho^\phi(\mathbf{x}_0)+\nabla f_\rho^\phi(\mathbf{x}_0)\cdot(\mathbf{x}-\mathbf{x}_0)$.
When $\mathbf{x}_j$ is close to $\mathbf{x}_i$,
$f_\rho^\phi(\mathbf{x}_j)\approx f^0(\mathbf{x}_i)+\vec{f}(\mathbf{x}_i)\cdot(\mathbf{x}_j-\mathbf{x}_i)$, where
$\vec{f}:=(f^1,\cdots,f^p)$ with $f^j=\partial f_\rho^\phi/\partial x^j$ for $j=1,\cdots,p$, and  $f^0$ is a new function introduced to approximate $f_\rho^\phi(\mathbf{x}_j)$.  The introduction of  $f^0$ is unavoidable since $y_j$ is valued $-1$ or $1$
and not a good approximation of $f_\rho^\phi$ at all.
After considering Taylor expansion between all pairs of samples, we define the following empirical error term for classification
\begin{equation}\label{eq fidelity term for classification}
\mathcal{E}^\phi_\mathcal{Z}(f^0,\vec{f}):=\frac{1}{n^2}\sum_{i,j=1}^n\omega_{i,j}^s\phi(y_j(f^0(\mathbf{x}_i)+\vec{f}(\mathbf{x}_i)
\cdot(\mathbf{x}_j-\mathbf{x}_i))),
\end{equation}
where $\omega_{i,j}^s$ is the weight function as in \eqref{eq weight}.

For the regularization term, we introduce
\begin{equation}
\Omega(f^0,\vec{f})=\lambda_1\|f^0\|_\mathcal{K}^2+\lambda_2\sum_{i=1}^p\|f^i\|_\mathcal{K}.
\end{equation}
Comparing with the regularization term for regression,  we have included an extra term $\lambda_1\|f^0\|_\mathcal{K}^2$ to control the smoothness of the $f^0$ function. We use two regularization parameters $\lambda_1$ and $\lambda_2$ for the trade-off between $\|f^0\|_\mathcal{K}^2$ and $\sum_{i=1}^p\|f^i\|_\mathcal{K}$.

Combining the data fidelity term and regularization term,  we formulate the sparse gradient learning for classification as follows
\begin{equation}\label{eq functional for classification}
(f_\mathcal{Z}^\phi,\vec{f}_\mathcal{Z}^\phi)=\arg\min_{(f^0,\vec{f})\in \mathbb{H}^{p+1}_\mathcal{K}}
\mathcal{E}_\mathcal{Z}^\phi(f^0,\vec{f})+\Omega(f^0,\vec{f}).
\end{equation}

\subsection{Forward-backward splitting for classification}
Using representer theorem, the minimizer of the infinite dimensional optimization problem in Eq.\ \eqref{eq functional for classification} has the following finite dimensional representation
\begin{equation*}
f^\phi_\mathcal{Z}=\sum_{i=1}^n\alpha_{i,\mathcal{Z}}\mathcal{K}(\mathbf{x},\mathbf{x}_i),\qquad (f^\phi_\mathcal{Z})^j=\sum_{i=1}^nc_{i,\mathcal{Z}}^j
\mathcal{K}(\mathbf{x},\mathbf{x}_i)
\end{equation*}
where $\alpha_{i,\mathcal{Z}},c_{i,\mathcal{Z}}^j\in \mathbb{R}$ for $i=1,\ldots,n$ and $j=1,\ldots,p.$

Then using the same technique as in the regression setting, the objective functional in minimization problem \eqref{eq functional for classification}
can be reformulated as a finite dimensional convex function of vector $\alpha=(\alpha_1,\ldots,\alpha_n)^T$ and
matrix $\widetilde{C}=(\widetilde{\mathbf{c}}_i^j)_{i=1,j=1}^{n,p}$. That is,
\begin{equation*}
\Psi(\alpha,\widetilde{C})=\frac{1}{n^2}\sum_{i,j=1}^n\omega_{i,j}^s\phi(y_j(\alpha^T\mathbf{k}_i+(\mathbf{x}_j-
\mathbf{x}_i)^T\widetilde{C}\mathbf{k}_i^{\frac{1}{2}}))+\lambda_1\alpha^TK\alpha+\lambda_2\sum_{j=1}^p\|\widetilde{\mathbf{c}}^j\|_2.
\end{equation*}
Then the corresponding finite dimensional convex
\begin{equation}\label{min Psi classification}
(\widetilde{\alpha}_\mathcal{Z}^\phi,\widetilde{C}_\mathcal{Z}^\phi)=\arg\min_{\alpha\in\mathbb{R}^n,
\widetilde{C}\in\mathbb{R}^{p\times n}}\Psi(\widetilde{C})
\end{equation}
can be solved by the forward-backward splitting algorithm.

We split $\Psi(\alpha,\widetilde{C})=\Psi_1+\Psi_2$ with $\Psi_1=\lambda_2\sum_{j=1}^p\|\widetilde{\mathbf{c}}^j\|_2$ and $\Psi_2=
\frac{1}{n^2}\sum_{i,j=1}^n\omega_{i,j}^s\phi(y_j(\alpha^T\mathbf{k}_i+(\mathbf{x}_j-
\mathbf{x}_i)^T\widetilde{C}\mathbf{k}_i^{\frac{1}{2}}))+\lambda_1\alpha^TK\alpha$.
Then the forward-backward splitting algorithm for solving \eqref{min Psi classification} becomes
\begin{equation}\label{iteration minimization classification}
\begin{cases}
\alpha^{(k+1)}=\alpha^{(k)}-\delta\left(\frac{1}{n^2}\sum_{i,j=1}^n\frac{-\omega_{ij}y_j\mathbf{k}_i}{1+\exp(y_j((\alpha^{(k)})^T\mathbf{k}_i+
(\mathbf{x}_j-\mathbf{x}_i)^T\widetilde{C}^{(k)}\mathbf{k}_i^{\frac{1}{2}}))}+2\lambda_1K\alpha^{(k)}\right),\cr
D^{(k+1)}=\widetilde{C}^{(k)}-
\frac{\delta U}{n^2}\sum_{i,j=1}^n\frac{-\omega_{i,j}^sy_j(\beta_j-\beta_i)(\mathbf{k}_i^{1/2})^T}{1+\exp(y_j((\alpha^{(k)})^T\mathbf{k}_i+(\mathbf{x}_j-\mathbf{x}_i)^T
\widetilde{C}^{(k)}\mathbf{k}_i^{\frac{1}{2}}))},
\cr
\widetilde{C}^{(k+1)}=T_{\lambda_2\delta}(D^{(k+1)}),
\end{cases}
\end{equation}
where $U,\beta$ satisfy equation \eqref{eq Sigmax decomposition} with $U$ being a $p\times n$ unitary matrix.

With the derived $\widetilde{C}_{\mathcal{Z}}^\phi$, we can do variable selection and
dimension reduction as done for the regression setting. We omit the details here.

\section{Examples}\label{sec
experiment}
Next we illustrate the effectiveness of variable selection and dimension reduction by sparse gradient learning
algorithm (SGL) on both artificial datasets and a gene expression dataset. As our method is a kernel-based method, known to be effective for nonlinear problems, we focus our experiments on nonlinear settings for the artificial datasets, although the method can be equally well applied to linear problems.

Before we report the detailed results, we would like to mention  that our forward-backward splitting algorithm is very efficient for solving the sparse gradient learning problem.
For the simulation studies, it takes only a few minutes to obtain the results to be described next. For the gene expression data involving 7129 variables, it takes less than two minutes to learn the optimal gradient functions on a modest desktop.

\subsection{Simulated data for regression}\label{subsec art regression}
In this example, we illustrate the utility of sparse gradient learning for variable selection by comparing it to the popular variable selection method LASSO.  We pointed out in section \ref{sec sparse gradient} that LASSO, assuming the prediction function is linear, can be viewed as a special case of sparse gradient learning. Because sparse gradient learning makes no assumption on the linearity of the prediction function, we expect it to be better equipped than LASSO for selecting variables with nonlinear responses.

We simulate $100$ observations from the model
\begin{equation*}
y=(2x^1-1)^2+x^2+x^3+x^4+x^5+\epsilon,
\end{equation*}
where $x^i, i=1,\ldots,5$ are i.i.d. drawn from uniform distribution on $[0,1]$ and $\epsilon$ is drawn form standard normal
distribution with variance $0.05$. Let $x^i, i=6,\ldots, 10$ be additional five noisy variables, which are also i.i.d. drawn
from uniform distribution on $[0,1]$.  We assume the observation dataset is given in the form of
$\mathcal{Z}:=\{\mathbf{x}_i,y_i\}_{i=1}^{100}$,
where $\mathbf{x}_i=(x_i^1,x_i^2,\ldots,x_i^{10})$ and $y_i=(2x_i^1-1)^2+x_i^2+x_i^3+x_i^4+x_i^5+\epsilon$. It is easy to see
that only the first $5$ variables contribute the value of $y$.

This is a well-known example as pointed out by B. Turlach in \citep{EHJT:AS:2004} to
show the deficiency of LASSO. As the ten variables are uncorrelated, LASSO will select variables based on their correlation
with the response variable $y$. However, because $(2x^1-1)^2$ is a symmetric function with respect to symmetric axis $x^1=\frac{1}{2}$ and
the variable $x^1$ is drawn from a uniform distribution on $[0,1]$, the correlation between $x^1$ and $y$
is $0$. Consequently, $x^1$ will not be selected by LASSO.  Because SGL selects variables based
on the norm of the gradient functions, it has no such a limitation.

To run the SGL algorithm in this example, we use the truncated Gaussian in Eq.\ \eqref{eq:truncated-weight} with $10$
neighbors as our weight function. The bandwidth parameter $s$ is chosen to be half of the median of the pairwise distances of
the sampling points.
As the gradients of the regression function with respect to different variables are all linear, we choose $\mathcal{K}(\mathbf{x},\mathbf{y})=1+\mathbf{x}\mathbf{y}$.

Figure 1 shows the variables selected by SGL and LASSO for the same dataset when the regularization parameter varies. Both
methods are able to successfully select the four linear variables (i.e. $x^2,\cdots,x^4$). However, LASSO failed to select $x^1$ and
treated $x^1$ as if it were one of five noisy term $x^6,\cdots, x^{10}$ (Fig. 1b). In contrast, SGL is clearly able to differentiate
$x^1$ from the group of five noisy variables (Fig. 1a).

To summarize how often each variable will be selected, we repeat the simulation $100$ times. For each simulation, we choose
a regularization parameter so that each algorithm returns exactly five variables. Table \ref{table regression} shows the frequencies of variables  $x^1,x^2,\ldots,x^{10}$ selected by SGL and LASSO in $100$ repeats. Both methods are able to select the four linear variables,
$x^2,x^3,x^4,x^5$, correctly. But, LASSO fails to select $x^1$ and treats it as the same as the noisy variables $x^6, x^7,x^8,x^9,x^{10}$.
This is in contrast to SGL, which is able to correctly select $x^1$ in 78\% of the times, much greater than the frequencies (median $5\%$)
of selecting the noisy variables. This example illustrates the advantage of SGL for variable selection in nonlinear settings.

\begin{table}
\begin{center}
\caption{\label{table regression}}{\small Frequencies of variables $x^1,x^2,\ldots,x^{10}$ selected by SGL and LASSO in
$100$ repeats}
\end{center}
\begin{center}
\begin{tabular*}{0.75\textwidth}{@{\extracolsep{\fill}} |l|c|c|c|c|c|c|c|c|c|c|}
\hline variable& $x^1$ & $x^2$ & $x^3$ & $x^4$ &$x^5$ &$x^6$ &$x^7$&$x^8$&$x^9$&$x^{10}$\\ \hline SGL
& $78$ & $100$ & $100$ & $100$&$100$&$7$&$4$&$6$&$5$&$2$ \\
\hline LASSO
& $16$ & $100$ & $100$ & $100$&$100$&$25$&$14$&$13$&$13$&$19$\\
\hline
\end{tabular*}
\end{center}
\end{table}

\begin{figure}
\subfigure[]{\includegraphics[width=.45\textwidth]{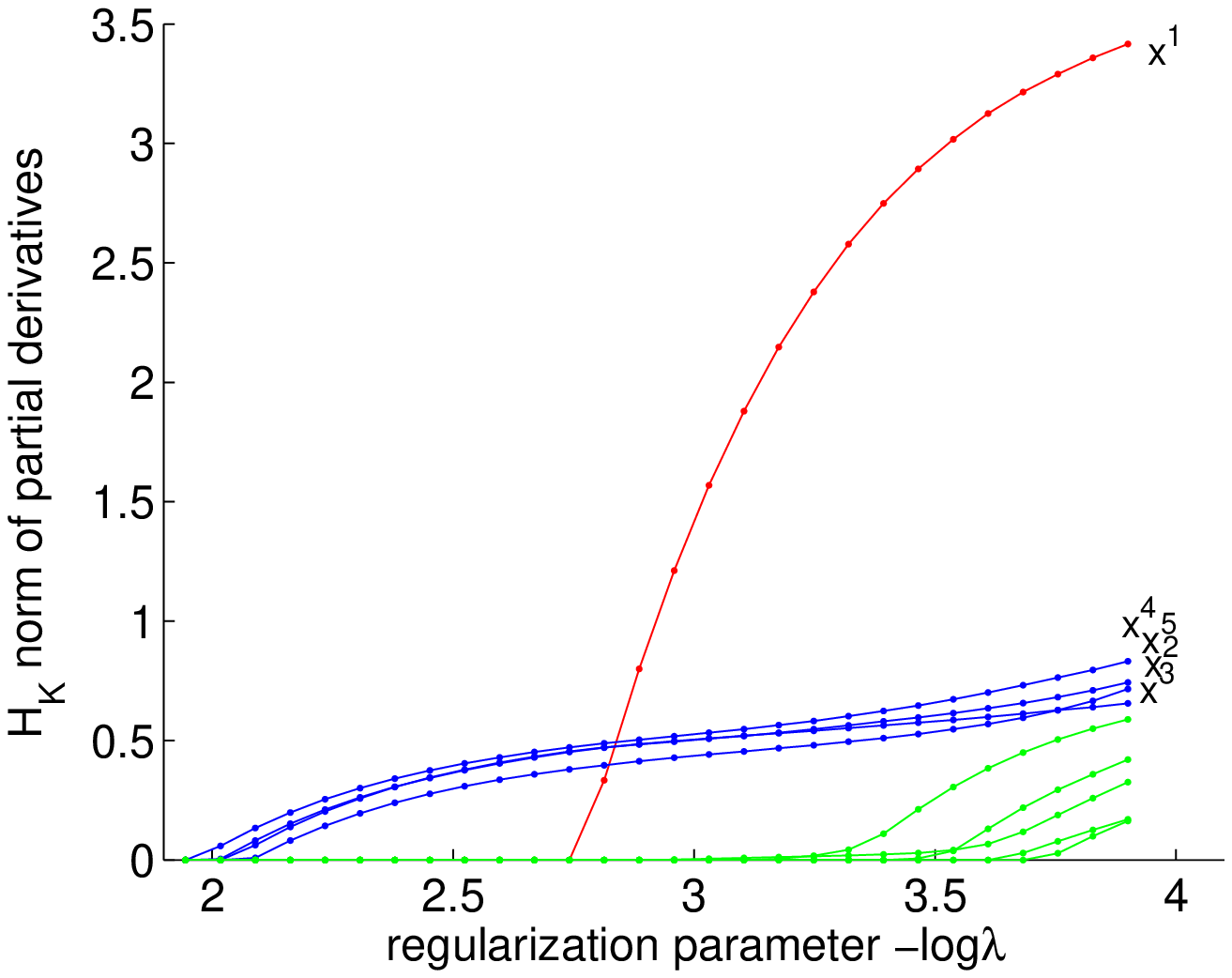}}
\subfigure[]{\includegraphics[width=.45\textwidth]{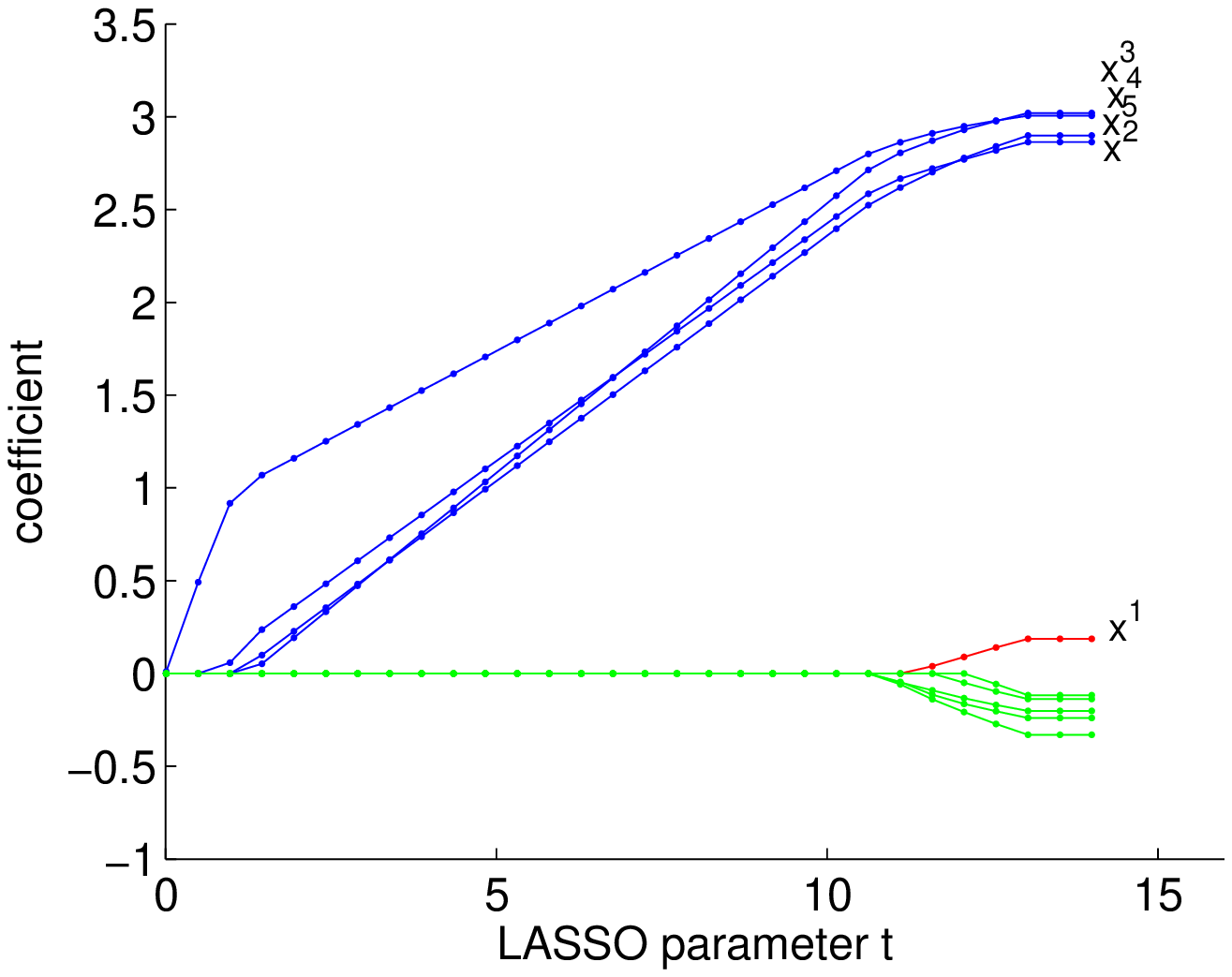}}
\caption{Regularization path for SGL and LASSO. Red line represents the variable $x^1$,
blue lines represent the variables $x^2,x^3,x^4,x^5$ and green lines represent noisy variables
$x^6,x^7,x^8,x^9,x^{10}$. (a)$H_K$ norm of each partial derivatives derived by SGL with respect to regularization parameter,
where regularization parameter is scaled to be $-\log\lambda$ with base $10$.
(b)LASSO shrinkage of coefficients with respect to LASSO parameter $t$.}\label{figure regularization path}
\end{figure}

\subsection{Simulated data for classification}
Next we apply SGL to an artificial dataset that has been commonly used to test the efficiency of dimension reduction methods
in the literature. We consider a binary classification problem in which the sample data are lying in a $200$ dimensional space
with only the first  $2$ dimensions being relevant for classification and the remaining variables being noises.
 More specifically, we generate $40$ samples with half from $+1$ class
and the other half from $-1$ class. For the samples from $+1$ class, the first $2$-dimensions of the sample data correspond to points drawn uniformly from  a $2$-dimensional spherical surface with radius $3$.  The remaining $198$ dimensions are
noisy variables with each variable being i.i.d drawn from Gaussian distribution $N(0,\sigma)$. That is,
\begin{equation}\label{art_noise variable distribution}
x^j\sim N(0,\sigma),\quad \hbox{for}\ j=3,4,\ldots,200.
\end{equation}
For the samples from $-1$ class, the first $2$-dimensions of the sample data correspond to points drawn uniformly from a $2$-dimensional
spherical surface with radius $3\times2.5$ and the remaining $198$ dimensions are noisy variables with each
variable $x^j$ i.i.d drawn from $N(0,\sigma)$ as \eqref{art_noise variable distribution}. Obviously, this data set can be
easily separated by a sphere surface if we project the data to the  Euclidean space spanned by the first two dimensions.

In what follows, we illustrate the effectiveness of SGL on this data set for both variable selection and dimension reduction.
In implementing SGL, both the weight function and the kernel are
all chosen to be $\exp(-\frac{\|\mathbf{x}-\mathbf{u}\|^2}{2s^2})$ with $s$ being half of the median of pairwise distance
of the sampling points.

\begin{figure}[htp]
\subfigure[]{\includegraphics[width=.30\textwidth]{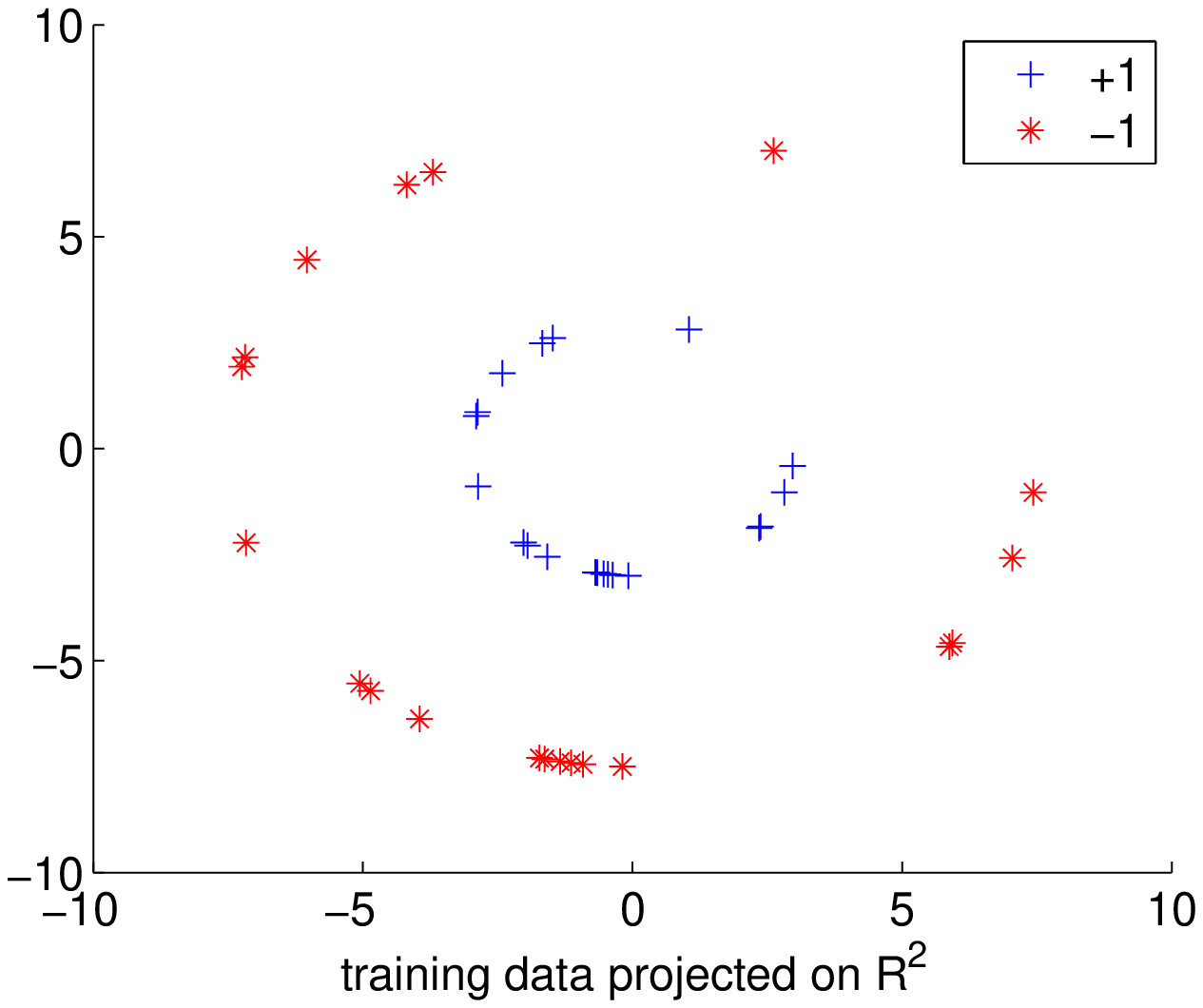}}
\subfigure[]{\includegraphics[width=.30\textwidth]{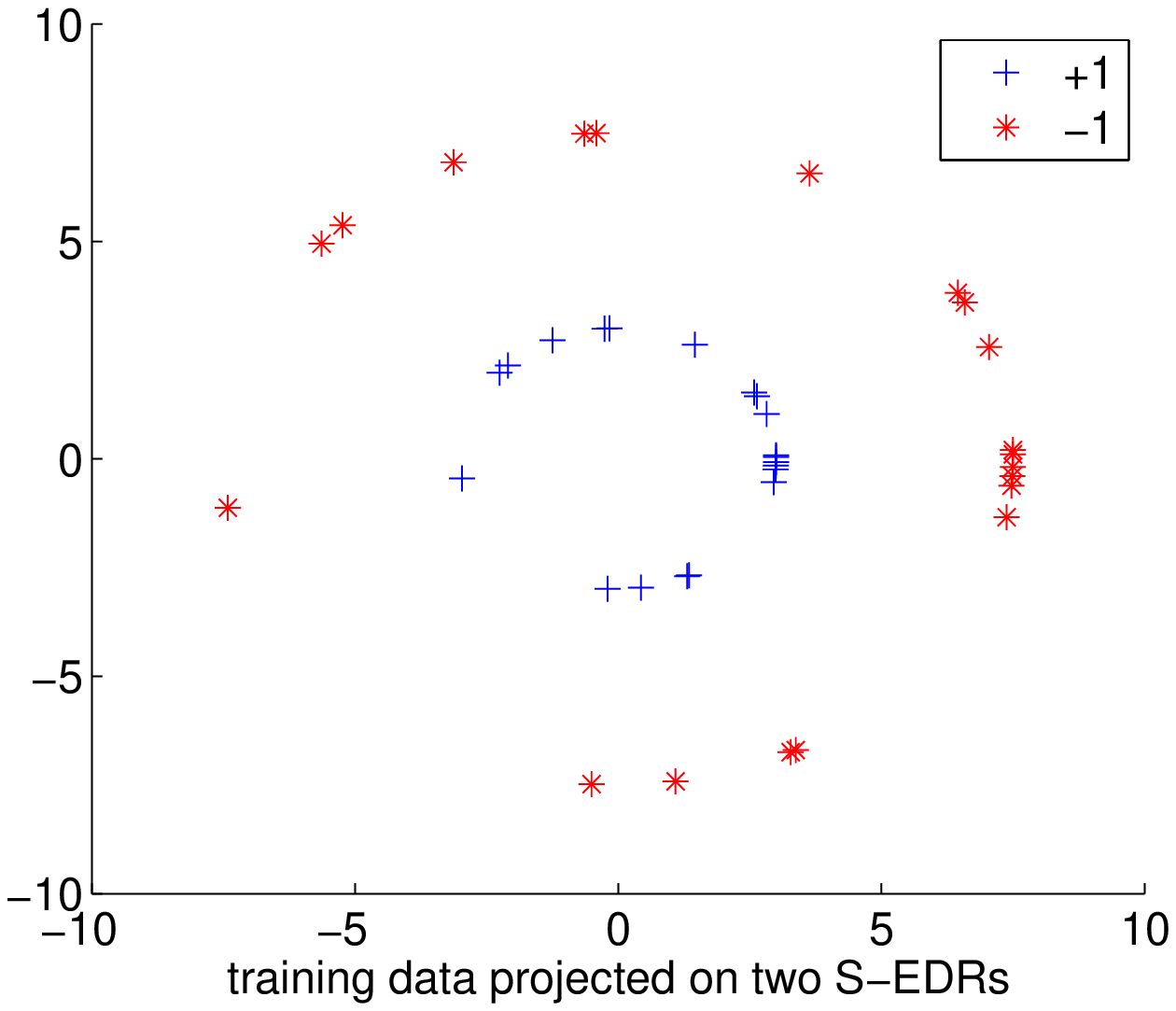}}
\subfigure[]{\includegraphics[width=.30\textwidth]{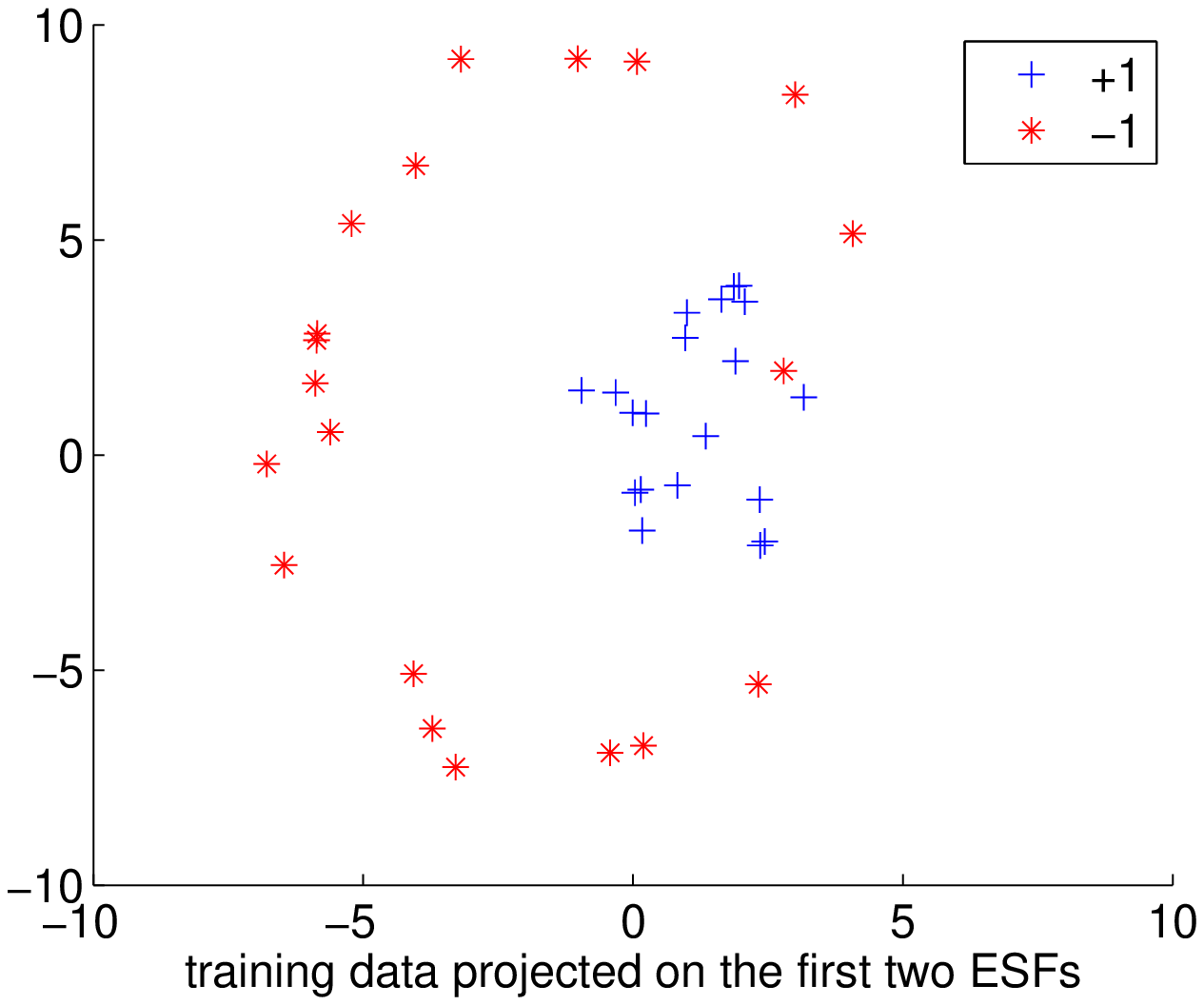}}\\
\subfigure[]{\includegraphics[width=.30\textwidth]{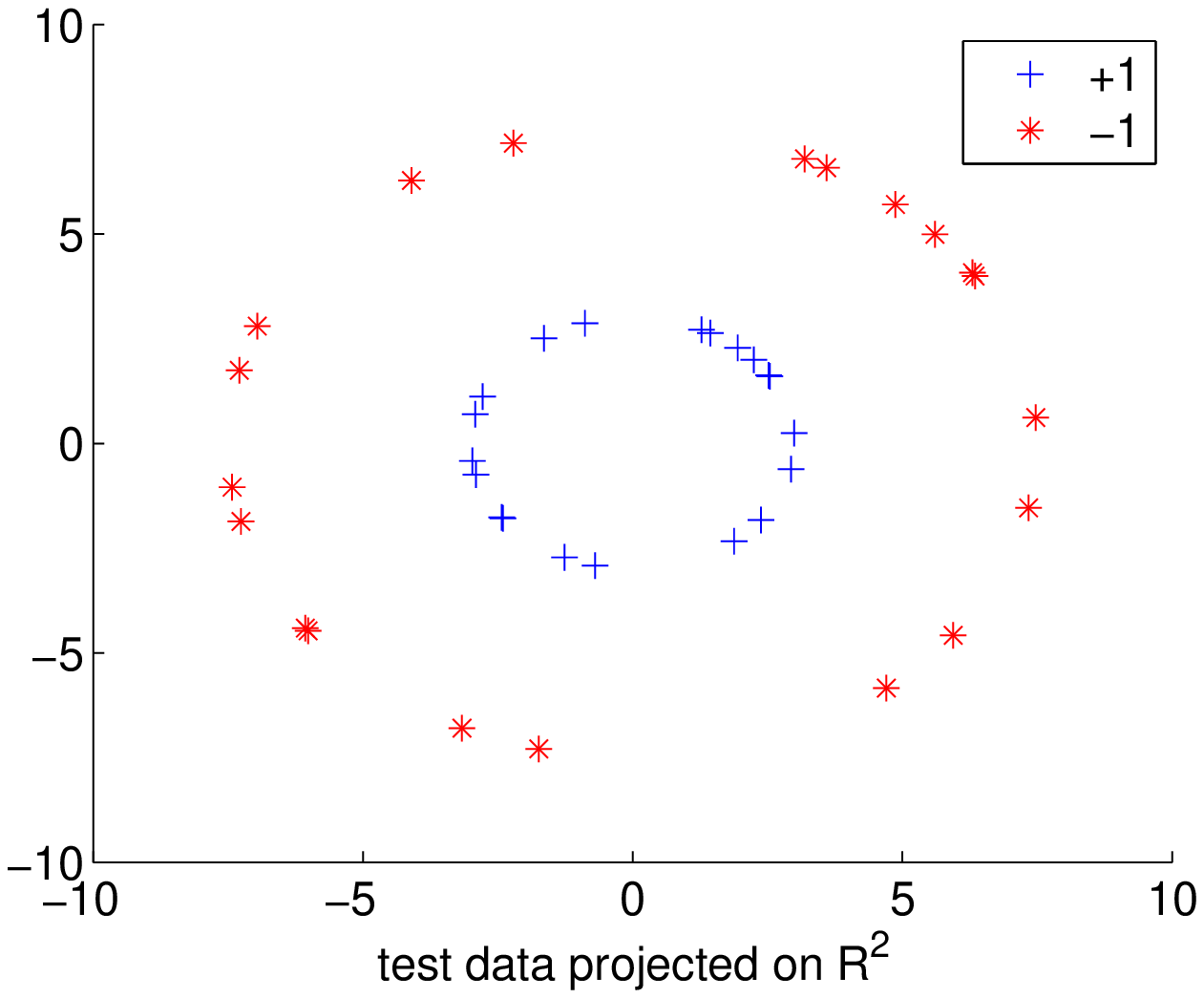}}
\subfigure[]{\includegraphics[width=.30\textwidth]{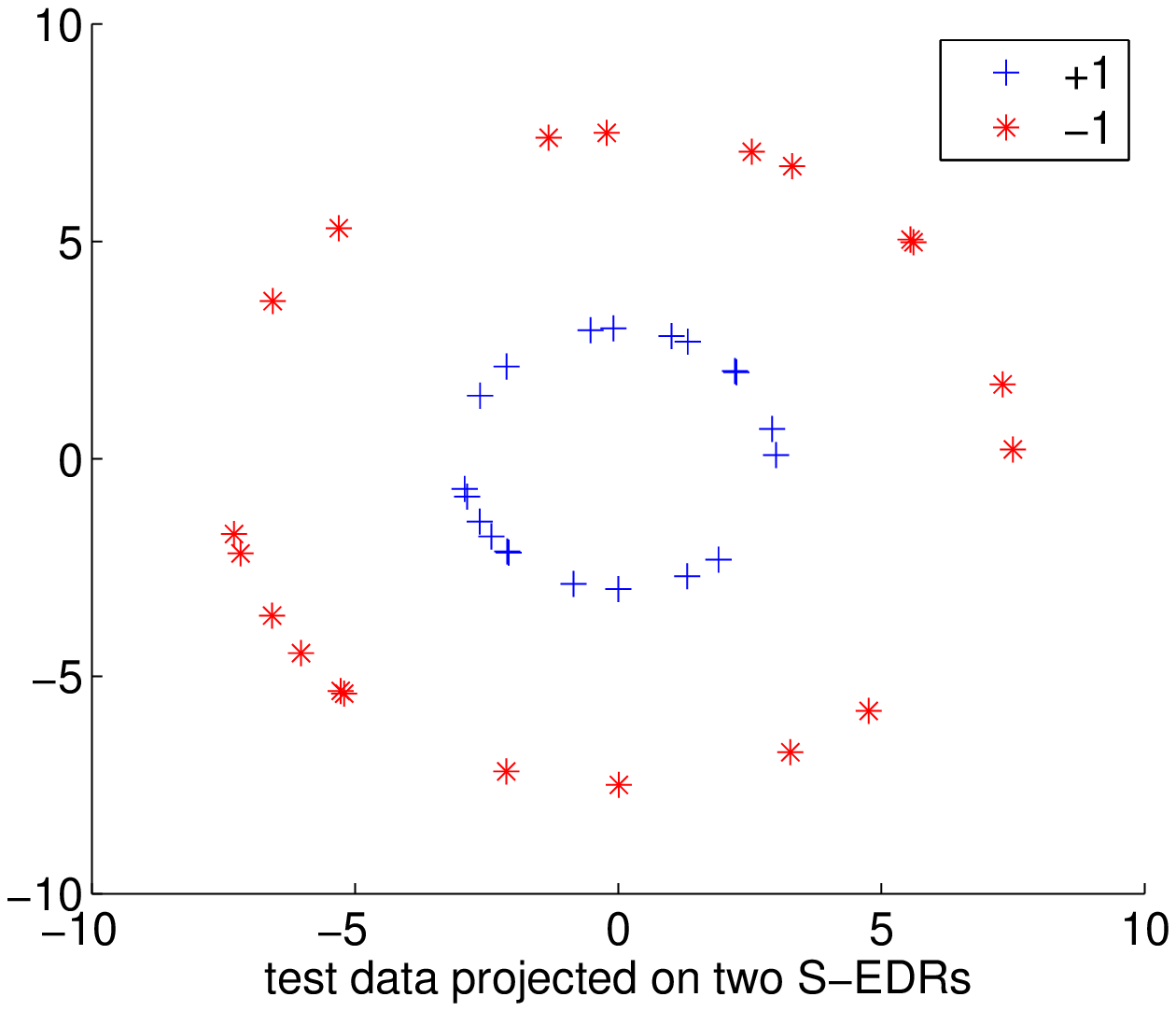}}
\subfigure[]{\includegraphics[width=.30\textwidth]{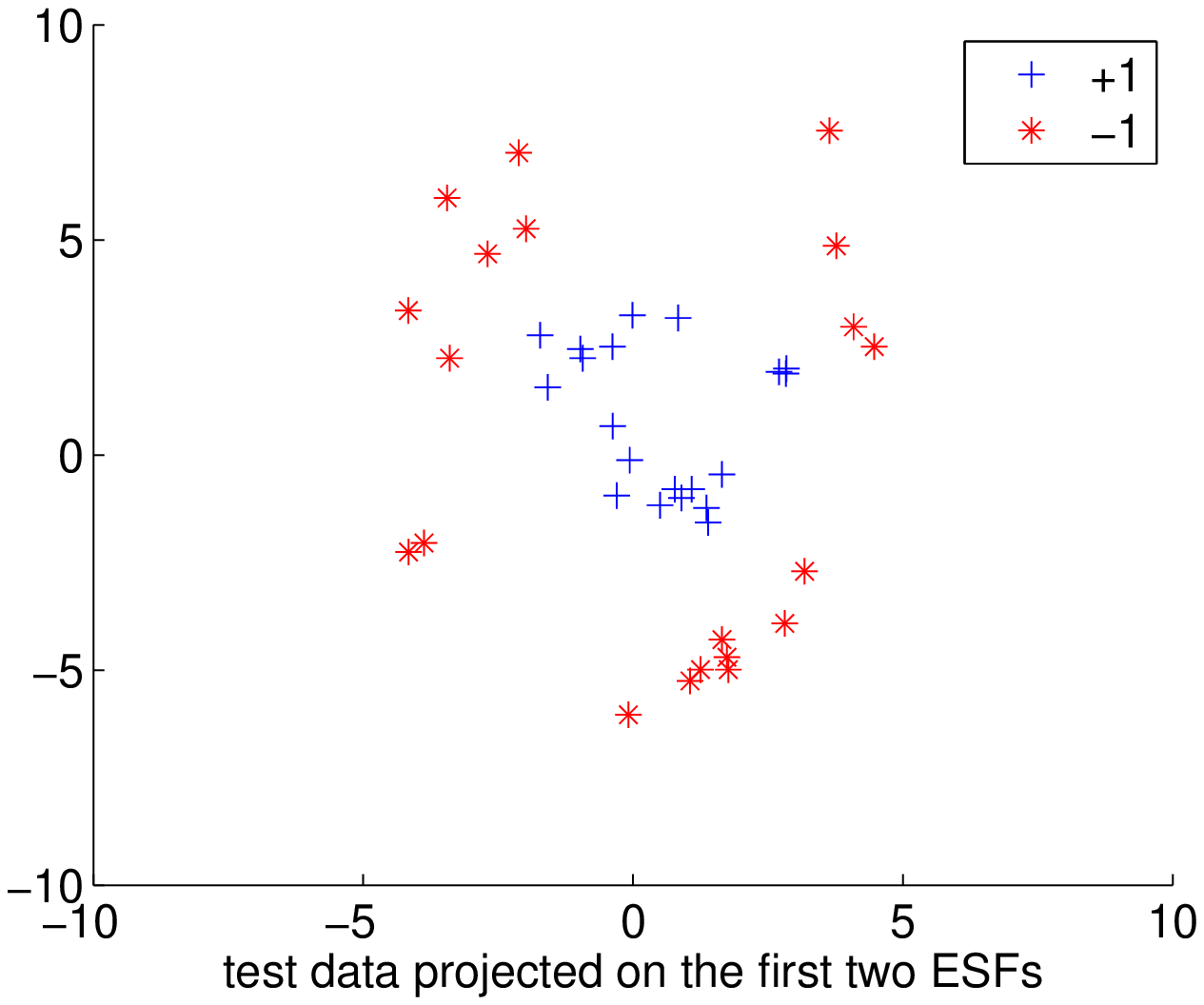}}
\caption{Nonlinear classification simulation with $\sigma=3$.  (a) Training data
projected on the first two dimensions, (b) Training data projected on two S-EDRs
derived by SGL.
(c)Training data projected on first two ESFs derived by GL.  (d) Test data
projected on the first two dimensions.  (e) Test data projected on two S-EDRs
derived by SGL.
(f) Test data projected on first two ESFs derived by GL.}\label{art_class_GL_figure1}
\end{figure}

\begin{figure}[htp]
\subfigure[]{\includegraphics[width=.30\textwidth]{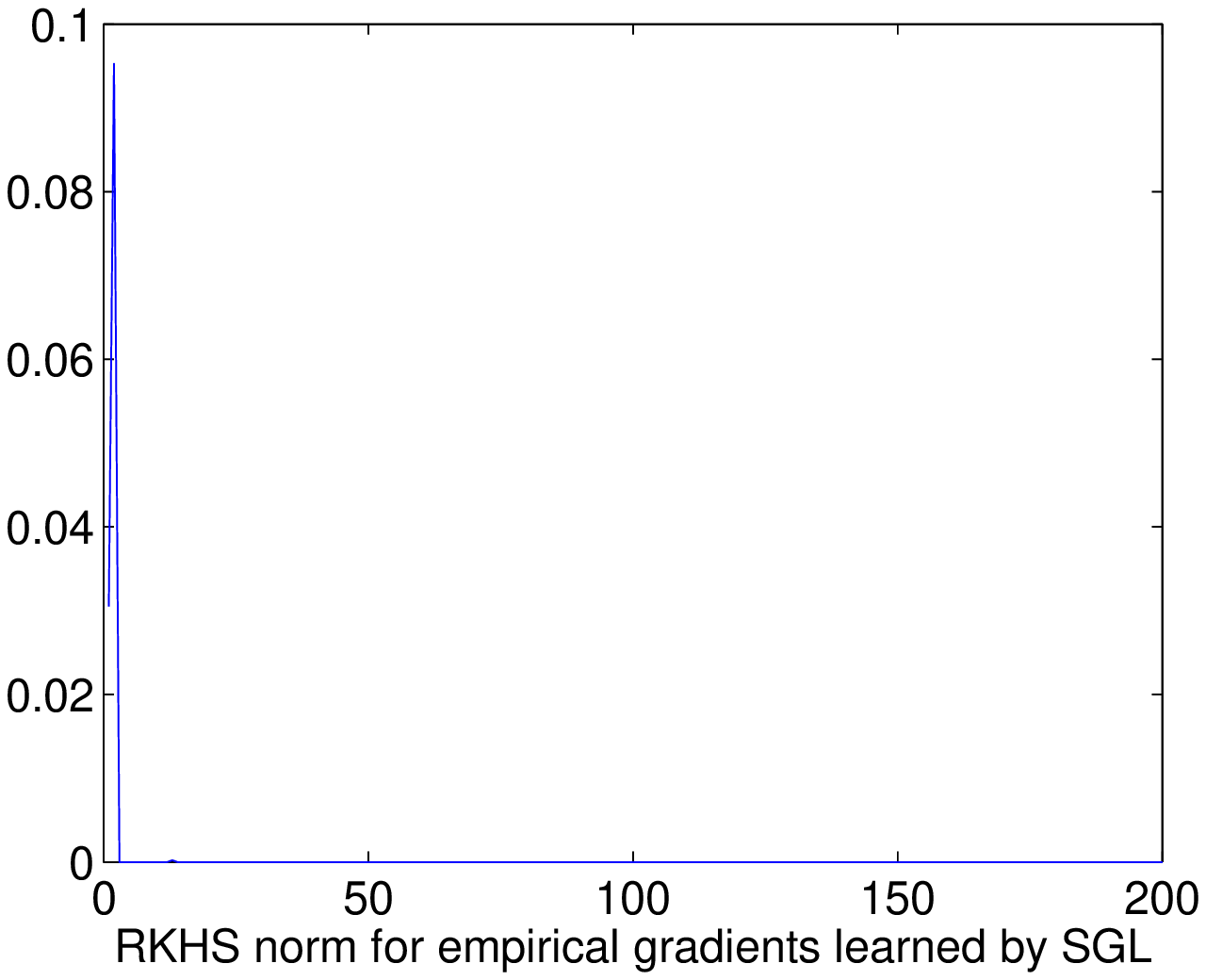}}
\subfigure[]{\includegraphics[width=.30\textwidth]{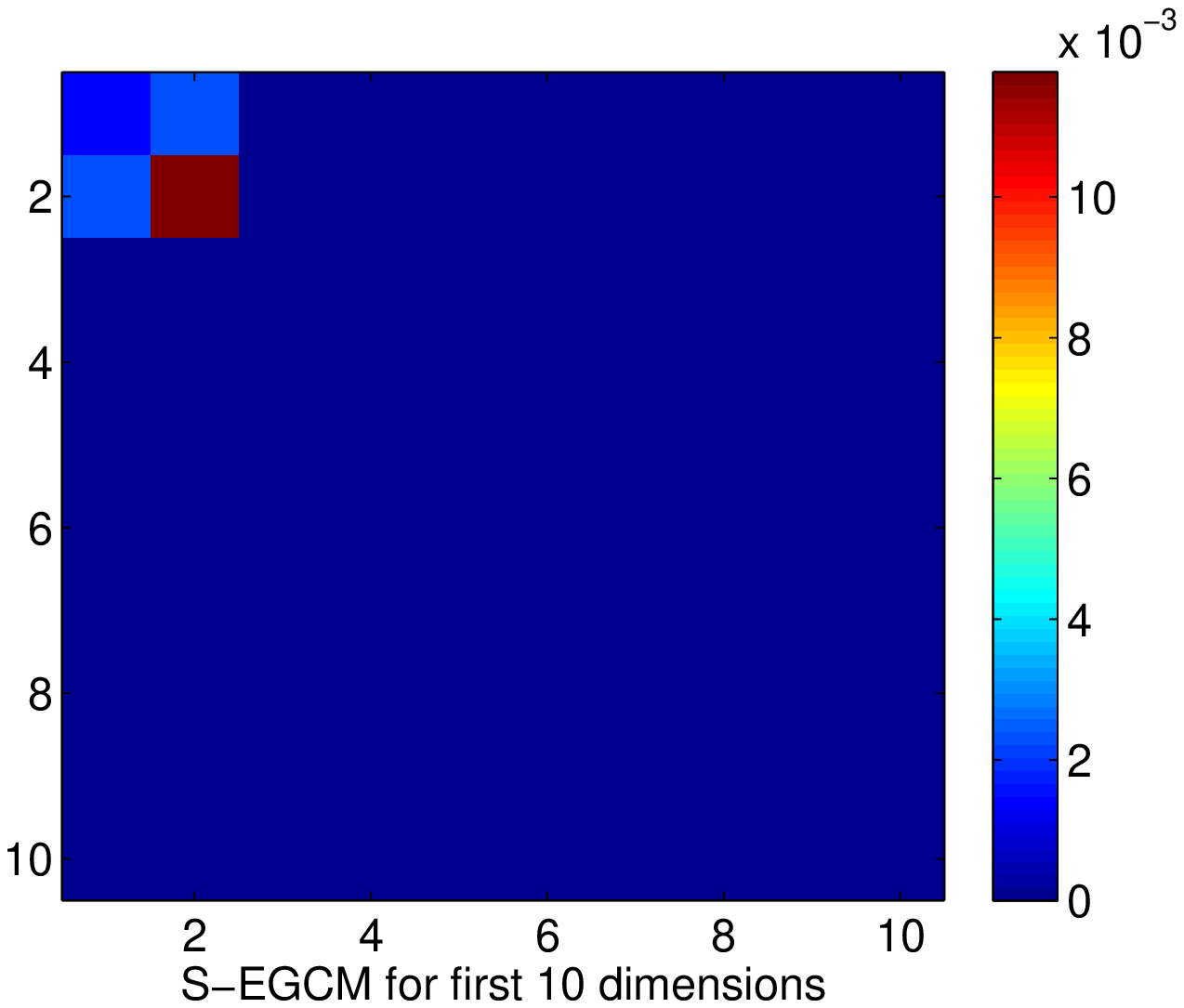}}
\subfigure[]{\includegraphics[width=.30\textwidth]{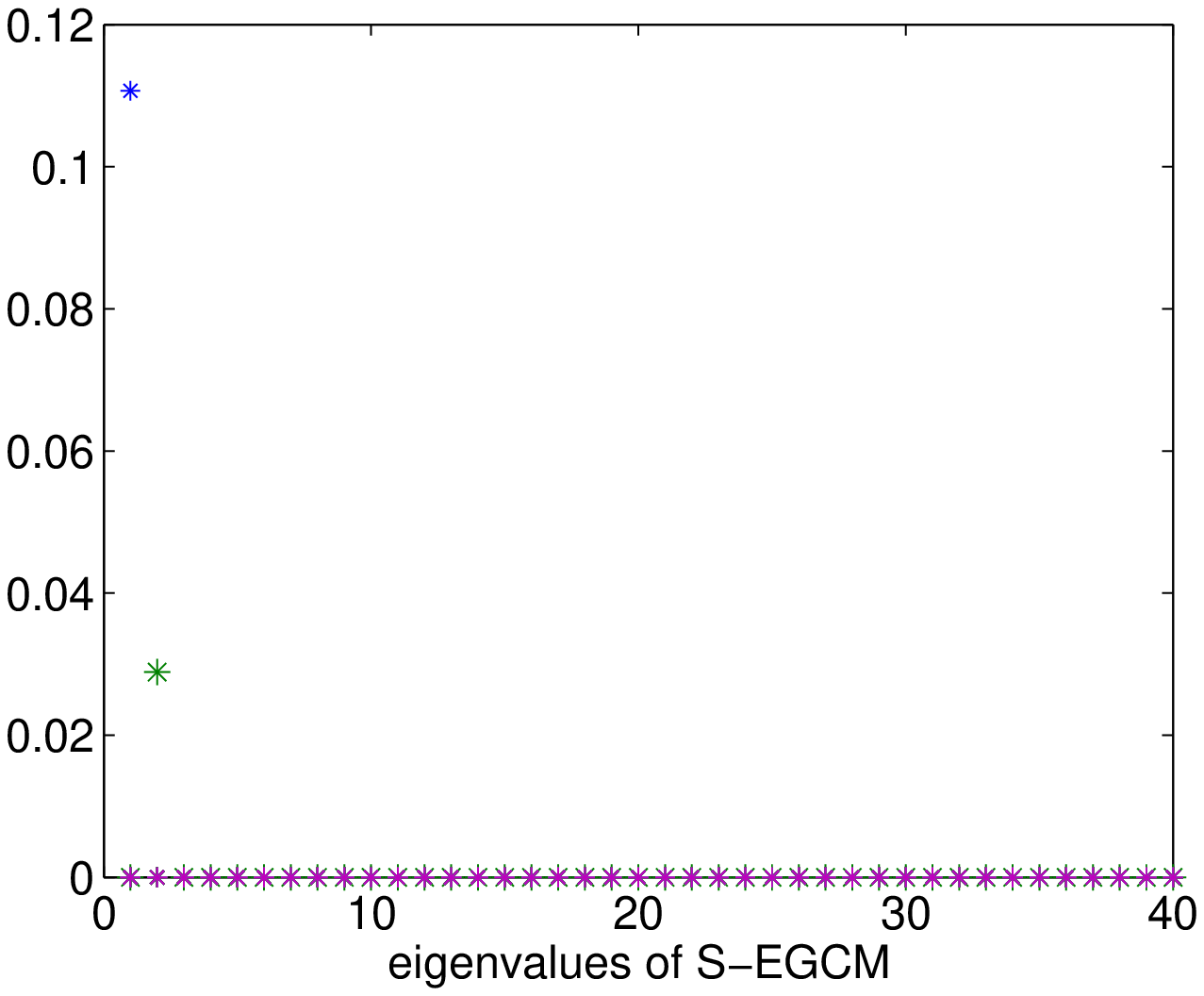}}\\
\subfigure[]{\includegraphics[width=.30\textwidth]{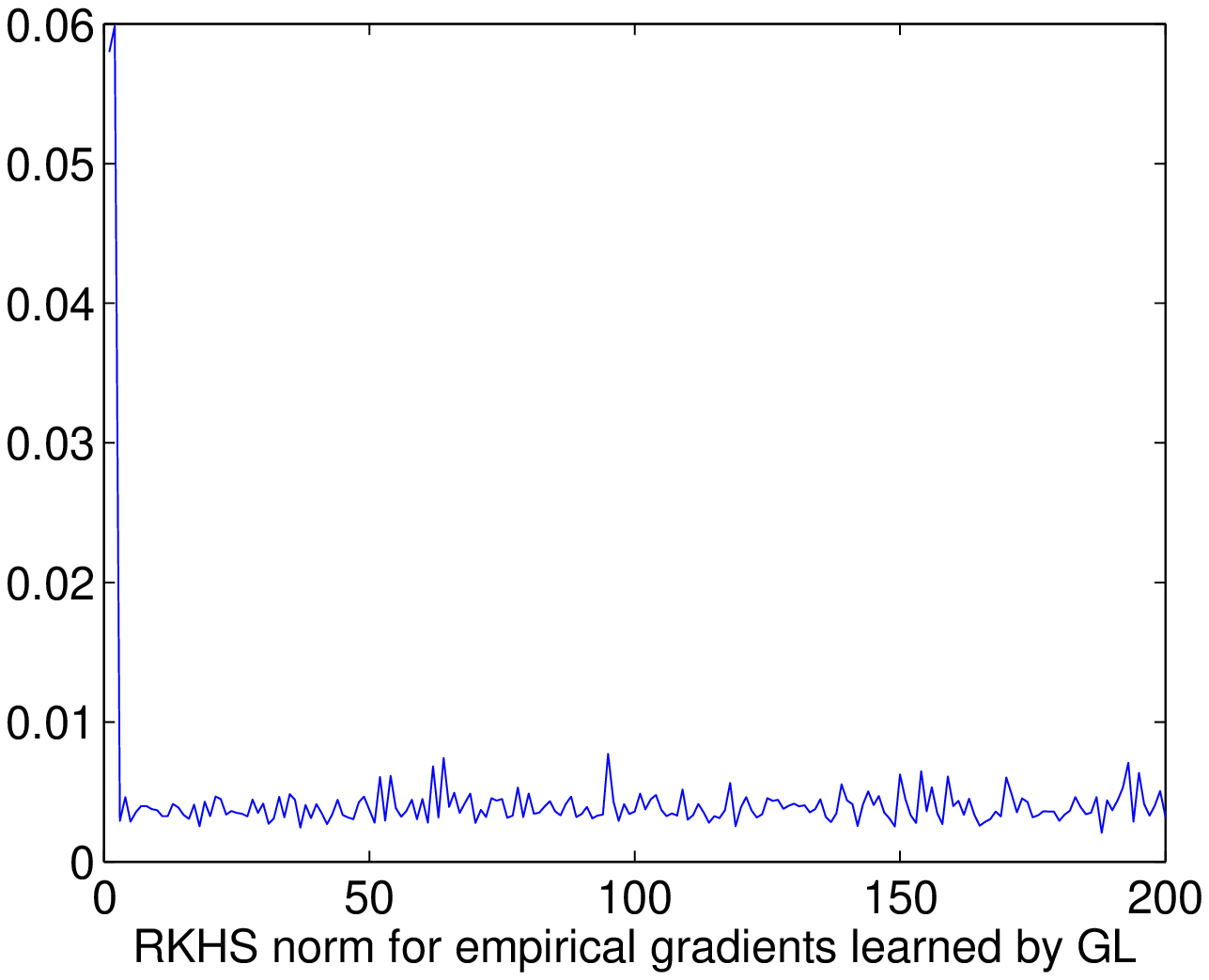}}
\subfigure[]{\includegraphics[width=.30\textwidth]{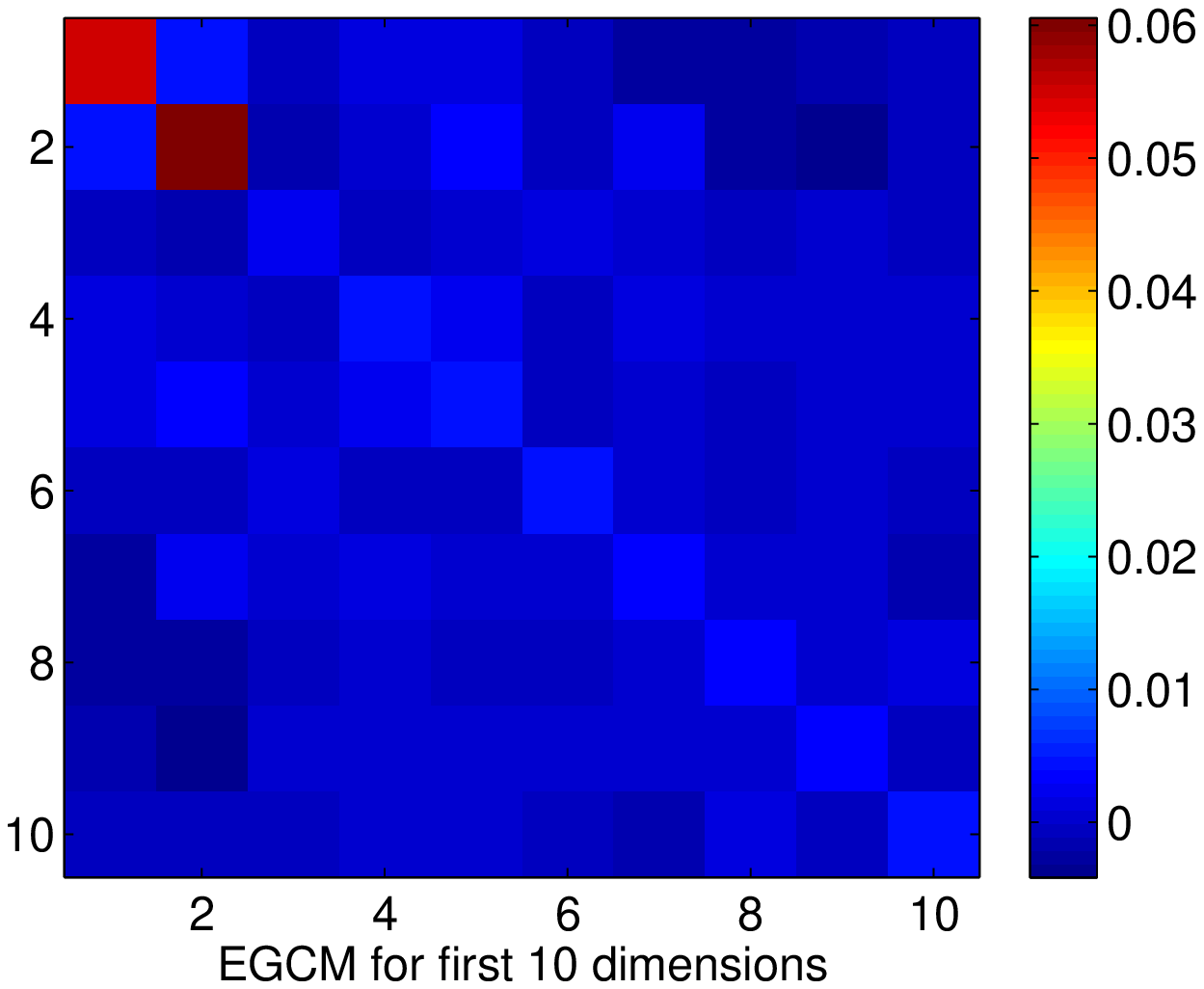}}
\subfigure[]{\includegraphics[width=.30\textwidth]{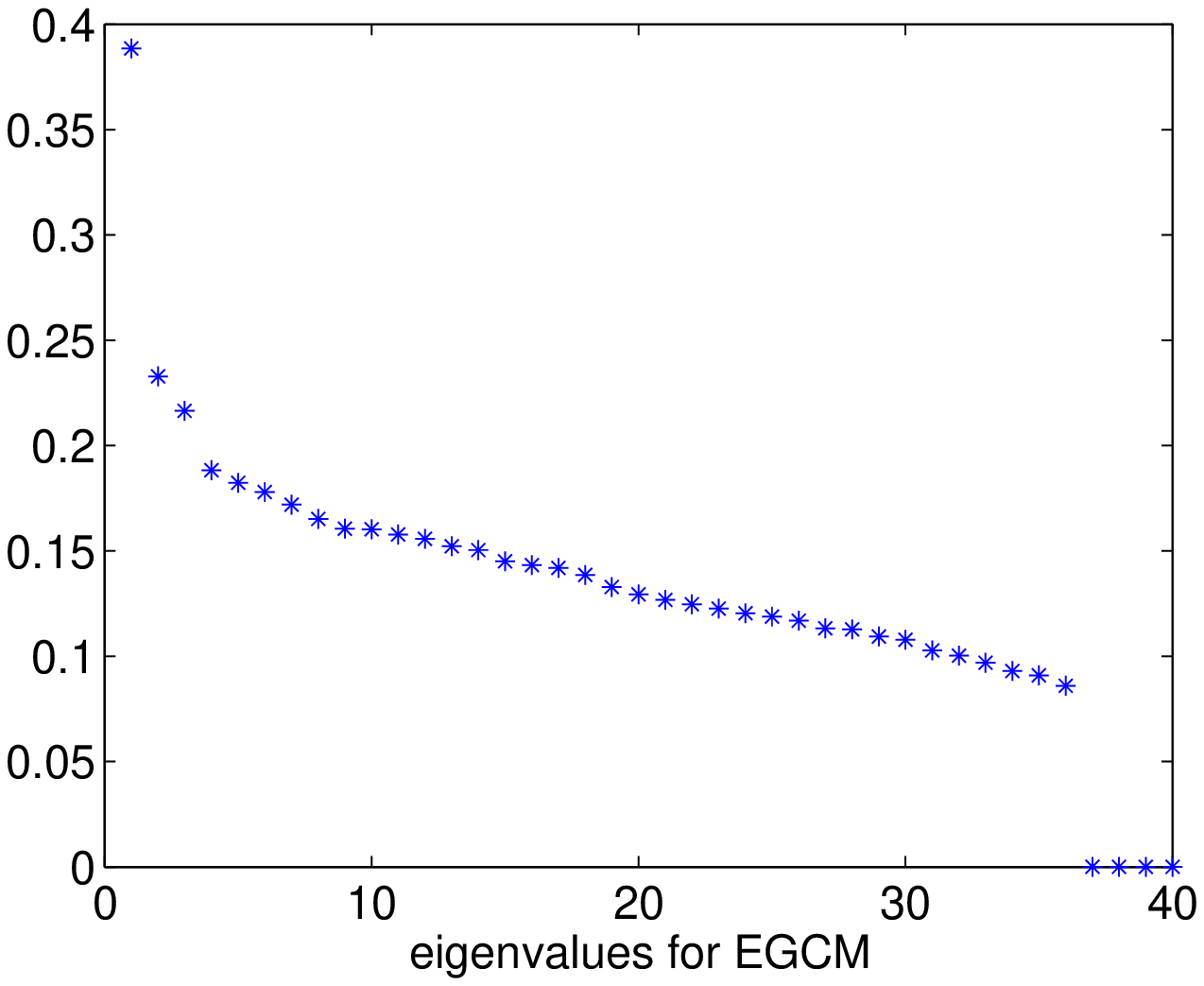}}
\caption{Nonlinear classification simulation with $\sigma=3$ (continued).  (a) RKHS norm of empirical gradient
derived by SGL. (b) S-EGCM for first 10 dimension.
(c) Eigenvalues of S-EGCM.  (d) RKHS norm of empirical gradient derived by GL,
(e) EGCM for first 10 dimension.
(f) Eigenvalues of EGCM.}\label{art_class_GL_figure2}
\end{figure}

We generated several datasets with different noise levels by varying $\sigma$ from $0.1$ to $3$.
SGL correctly selected $x^1$ and $x^2$ as the important variables for all cases we tested. Furthermore,
SGL also generated two S-EDRs that captured the underlying data structure for all these cases (Figure 2).
It is important to emphasize that the two S-EDRs generated by SGL are the only two features the algorithm can possibly
obtain, since the derived S-EGCM are supported on a $2\times 2$ matrix. As a result, both of the derived S-EDRs are
linear combinations of the first two variables. By contrast, using the gradient learning method (GL) reported in \citep{MWZ:Bernoulli:2009},
the first two returned dimension reduction directions (called ESFs) are shown to be able to capture the correct
underlying structure only when $\sigma<0.7$. In addition, the derived ESFs are linear combinations of all $200$
original variables instead of only two variables as in S-EDRs.
Figure \ref{art_class_GL_figure1}(b,e) shows the training data
and the test data projected on the derived two S-EDRs for a dataset with large noise ($\sigma=3$). Comparing to the
data projected on the first two dimensions (Figure \ref{art_class_GL_figure1}(a)(d)), the derived S-EDRs
preserves the structure of the original data.  In contrast, the
 gradient learning algorithm without sparsity constraint performed much poorer (Figure \ref{art_class_GL_figure1}(c)(f)).

To explain why SGL performed better than GL without sparsity constraint, we plotted
the norms of the derived empirical gradients from both methods in Figure \ref{art_class_GL_figure2}.
Note that although the norms of partial derivatives of unimportant variables derived from the method without sparsity constraint
are small, they are not exactly zero. As a result, all variables contributed and, consequently, introduced noise to
the empirical gradient covariance matrix (Figure \ref{art_class_GL_figure2}(e)(f)).

We also tested LASSO for this artificial data set, and not surprisingly it failed to identify the right
variables in all cases we tested.
We omit the details here.

\subsection{Leukemia classification}

Next we apply SGL to do variable selection and dimension reduction on gene expression data.
A gene expression data typically consists of the expression values of tens of thousands of mRNAs from
 a small number of samples as measured by microarrays.  Because of the large number of genes involved, the variable selection step
 becomes especially important both for the purpose of generating better prediction models, and also for elucidating biological
 mechanisms underlying the data.

 The gene expression data we will use is a widely studied dataset, consisting of the measurements of
  $7129$ genes from $72$ acute leukemia samples \citep{GSTPHMCLDC:Science:1999}. The samples are labeled with two leukemia types according to the precursor
  of the tumor cells - one is called acute lymphoblastic leukemia (ALL), and the other one is called acute myelogenous leukemia (AML). The two tumor types
  are difficult to distinguish morphologically, and the gene expression data is used to build a classifier to
  classify these two types.

Among $72$ samples, $38$ are training data and $34$ are test data.
We coded the type of leukaemia as a binary response variable $y$, with $1$ and $-1$ representing ALL and AML respectively.
The variables in the training samples
$\{\mathbf{x}_i\}_{i=1}^{38}$ are normalized to be zero mean and unit length for each gene.
The test data are similarly normalized, but only using the empirical mean and variance of the training data.

We applied three methods (SGL, GL and LASSO) to the dataset to select variables and extract the dimension reduction directions.
To compare the performance of the three methods, we used linear SVM to build a classifier based on the variables or features
returned by each method, and evaluated the classification performance using both leave-one-out (LOO) error on the training data
and the testing error. To implement SGL, the bandwidth parameter $s$ is chosen to be half of the median of the
pairwise distances of the sampling points, and $\mathcal{K}(\mathbf{x},\mathbf{y})=\mathbf{x}\mathbf{y}$.
The regularization parameters for the three methods are all chosen according to their prediction power
measured by leave-one-out error.

\begin{table}[htp]
\begin{center}\caption{\label{testerrortable_all_aml}}{\small Summary of the Leukemia classification results}
\end{center}
\begin{center}
\begin{tabular}{|l|c|c|c|c|c|}
\hline Method& SGL(variable selection) & SGL(S-EDRs) & GL(ESFs) & Linear SVM &LASSO \\ \hline number of variables or features
& $106$ & $1$ & $6$ & $7129$(all)&33 \\
\hline leave one out error (LOO)  &$0/38$& $0/38$& $0/38$& $3/38$&$1/38$\\
\hline test errors &$0/34$& $0/34$& $2/34$& $2/34$&$1/34$\\
\hline
\end{tabular}\end{center}
\end{table}

Table \ref{testerrortable_all_aml} shows the results of the three methods.  We implemented two SVM classifiers for SGL using either only
the variables or the features returned by SGL. Both classifiers are able to achieve perfect classification for both leave-one-out
and testing samples. The performance of SGL is better than both GL and LASSO, although only slightly. All three methods performed
significantly better than the SVM classifier built directly from the raw data.

In addition to the differences in prediction performance, we note a few other observations.
First,  SGL selects more genes than LASSO, which likely reflects
the failure of LASSO to choose genes with nonlinear relationships with the response variable,
as we illustrated in our first example. Second, The S-EDRs derived by SGL are linear
combinations of $106$ selected variables rather than all original variables as in the case of ESFs derived by GL.
This is a desirable property since an important goal of the gene expression analysis is to identify regulatory pathways
underlying the data, e.g. those distinguishing the two types of tumors. By associating only a small number of genes,
S-EDRs  provide better and more manageable candidate pathways for further experimental testing.

\section{Discussion}

Variable selection and dimension reduction are two common strategies for high-dimensional data analysis. Although many methods
have been proposed before for variable selection or dimension reduction,  few methods are currently available
for simultaneous variable selection and dimension reduction. In this work, we described a sparse gradient learning
algorithm that integrates automatic variable selection and dimension reduction into the same optimization framework.
The algorithm can be viewed as a generalization of LASSO from linear to non-linear variable selection, and a generalization of
the OPG method for learning EDR directions from a non-regularized to regularized estimation.  We showed that the integrated
framework offers several advantages over the previous methods by using both simulated and real-world examples.

The SGL method can be refined by using an adaptive weight function rather than a fixed one as in our current implementation.
The weight function $\omega_{i,j}^s$  is used to measure the distance between two sample points.  If the data are lying in
a lower dimensional space, the distance would be more accurately captured by using only variables related to the lower dimensional space
rather than all variables.  One way to implement this is to calculate the distance using only selected variables. Note that the forward-backward splitting algorithm eliminates variables at each step of the iteration. We can thus use an adaptive weight function that calculates the distances
based only on selected  variables returned after each iteration.  More specifically, let $\mathcal{S}^{(k)}=\{i:\|(\widetilde{\mathbf{c}}^{i})^{(k)}\|_2\neq0\}$ represent the variables selected after iteration $k$. An adaptive approach is to use $\sum_{l \in \mathcal{S}^{(k)}} (x_i^l - x_j^l)^2$ to measure
the distance $\|\mathbf{x}_i -\mathbf{x}_j\|^2$ after iteration $k$.

An interesting area for future research is to extend SGL for  semi-supervised learning.  In many applications, it is often much easier
to obtain unlabeled data with a larger sample size $u>>n$.  Most natural (human or animal)
learning seems to occur in semi-supervised settings \citep{BNS:JMLR:2006}.  It is possible to extend SGL for the semi-supervised learning
along several directions.  One way is to use the
unlabeled data
$\mathcal{X}=\{\mathbf{x}_i\}_{i=n+1}^{n+u}$ to control the approximate norm
of $\vec{f}$ in some Sobolev spaces and introduce a semi-supervised
learning algorithm as
\begin{eqnarray}
\vec{f}_{\mathcal{Z},\mathcal{X},\lambda,\mu}&=&
\arg\min_{\vec{f}\in {\cal
H}_K^p}\bigg\{\frac{1}{n^2}\sum_{i,j=1}^n\omega_{i,j}^s\big(y_i-y_j+\vec{f}(\mathbf{x}_i)\cdot
(\mathbf{x}_j-\mathbf{x}_i)\big)^2\nonumber\\&&+\frac{\mu}{(n+u)^2}\sum_{i,j=1}^{n+u}W_{i,j}\|\vec{f}(\mathbf{x}_i)-\vec{f}(\mathbf{x}_j)\|^2_{\ell^2(\mathbb{R}^p)}+
\lambda\|\vec{f}\|_K\bigg\},\nonumber
\end{eqnarray}
where $\|\vec{f}\|_K=\sum _{i=1}^p\|f^i\|_K$, ${W_{i,j}}$ are edge
weights in the data adjacency graph, $\mu$ is another regularization
parameter and often satisfies $\lambda=o(\mu)$. In order to make the algorithm efficiency,
we can use truncated weight in implementation as done in section \ref{subsec art regression}.

The regularization term $\sum_{i,j=1}^{n+u}W_{i,j}\|\vec{f}(\mathbf{x}_i)-\vec{f}(\mathbf{x}_j)\|^2_{\ell^2(\mathbb{R}^p)}$ is mainly motivated by the recent work of M. Belkin and P. Niyogi \citep{BNS:JMLR:2006}.
In that paper, they have introduced a regularization term $\sum_{i,j=1}^{n+u}W_{i,j}(f(\mathbf{x}_i)-f(\mathbf{x}_j))^2$ for  semi-supervised regression and classification problems. The term $\sum_{i,j=1}^{n+u}W_{i,j}(f(\mathbf{x}_i)-f(\mathbf{x}_j))^2$ is well-known to be related to graph Laplacian operator. It is used to approximate $\int_{\mathbf{x}\in \mathcal{M}}\|\nabla _\mathcal{M}f\|^2d\rho_X(\mathbf{x})$, where $\mathcal{M}$ is a compact submanifold which is the support of  marginal distribution $\rho_X(\mathbf{x})$, and $\nabla _\mathcal{M}$ is the
gradient of $f$ defined on $\mathcal{M}$ \citep{Carmo:book:1992}.  Intuitively, $\int_{\mathbf{x}\in \mathcal{M}}\|\nabla _\mathcal{M}f\|^2d\rho_X(\mathbf{x})$ is a smoothness penalty corresponding to the probability distribution. The idea behind  $\int_{\mathbf{x}\in \mathcal{M}}\|\nabla _\mathcal{M}f\|^2d\rho_X(\mathbf{x})$ is that it reflects the intrinsic structure of $\rho_X(\mathbf{x})$. Our regularization term $\sum_{i,j=1}^{n+u}W_{i,j}\|\vec{f}(\mathbf{x}_i)-\vec{f}(\mathbf{x}_j)\|^2_{\ell^2(\mathbb{R}^p)}$ is a corresponding vector form of $\sum_{i,j=1}^{n+u}W_{i,j}(f(\mathbf{x}_i)-f(\mathbf{x}_j))^2$ in \citep{BNS:JMLR:2006}. The regularization framework of the SGL for semi-supervised learning can thus be viewed as a generalization of  this previous work.


\bibliographystyle{unsrtnat}
\bibliography{ygb}

\end{document}